\documentclass[twoside]{article}

\usepackage[accepted]{aistats2026}

\usepackage[utf8]{inputenc}
\usepackage[T1]{fontenc}
\usepackage{natbib}
\setcitestyle{authoryear,round,semicolon}
\usepackage[colorlinks=true, allcolors=blue]{hyperref}
\usepackage{url}
\usepackage{booktabs}
\usepackage{amsfonts}
\usepackage{nicefrac}
\usepackage[protrusion=false]{microtype}
\usepackage{xcolor}
\usepackage{tikz}
\usepackage{graphicx}
\usepackage{subcaption}
\usepackage{algorithm}
\usepackage{algpseudocode}
\usepackage{amsthm}
\usepackage{float}
\usepackage{placeins}
\usepackage{multirow}
\usepackage{caption}
\usepackage{amsmath,amssymb}
\usepackage{subcaption}
\usepackage{listings}
\usepackage{stfloats}
\usepackage{xurl}
\usepackage{amsthm}
\usepackage{wrapfig}

\newtheorem{Definition}{Definition}
\newtheorem{Theorem}{Theorem}
\newtheorem{Lemma}{Lemma}
\newtheorem*{Proof}{Proof}

\newtheorem{Proposition}{Proposition}
\newtheorem{Corollary}{Corollary}

\usepackage{cleveref}
\crefname{Theorem}{Theorem}{Theorems}
\crefname{Lemma}{Lemma}{Lemmas}
\crefname{Definition}{Definition}{Definitions}
\crefname{Assumption}{Assumption}{Assumptions}
\crefname{Proposition}{Proposition}{Propositions}
\crefname{Corollary}{Corollary}{Corollaries}
\crefname{algorithm}{Algorithm}{Algorithms}
\crefname{appendix}{Appendix}{Appendices}

\def\bE{\mathbb{E}}

\DeclareRobustCommand{\l}{\left}
\DeclareRobustCommand{\r}{\right}

\begin{document}

\twocolumn[

\aistatstitle{ConMeZO: Adaptive Descent-Direction Sampling for Gradient-Free Finetuning of Large Language Models}

\aistatsauthor{ Lejs Deen Behric \qquad Liang Zhang \qquad Bingcong Li \qquad Kiran Koshy Thekumparampil }

\aistatsaddress{
\hspace{2.5em} Department of Computer Science, ETH Zurich \\
\hspace{2.5em} \texttt{\{lejs.behric, liang.zhang, bingcong.li\}@inf.ethz.ch}
\And
\hspace{7em} Amazon AGI Labs \\
\hspace{7em} \texttt{kkt@amazon.com}
} 

\runningtitle{ConMeZO: Adaptive Descent-Direction Sampling for Gradient-Free Finetuning of LLMs}

\runningauthor{Lejs Deen Behric, Liang Zhang, Bingcong Li, Kiran Koshy Thekumparampil}

]

\begin{abstract}
Zeroth‑order or derivative-free optimization (MeZO) is an attractive strategy for finetuning large language models (LLMs) because it eliminates the memory overhead of backpropagation. However, it converges slowly due to the inherent curse of dimensionality when searching for descent directions in the high-dimensional parameter space of billion-scale LLMs. We propose ConMeZO, a novel zeroth‑order optimizer that accelerates convergence by adaptive directional sampling. Instead of drawing the direction uniformly at random, ConMeZO restricts the sampling to a cone centered around a momentum estimate. This concentrates the search in directions where the true gradient is more likely to lie and thus reduces the effect of high dimensions. We prove that ConMeZO achieves the same worst-case convergence rate as MeZO. Empirically, when finetuning LLMs on natural language tasks, ConMeZO is up to 2$\times$ faster than MeZO while retaining the low‑memory footprint of zeroth-order methods.
\end{abstract}

\section{INTRODUCTION}

\begin{figure}[tb]
  \centering
    \includegraphics[width=0.8\linewidth]{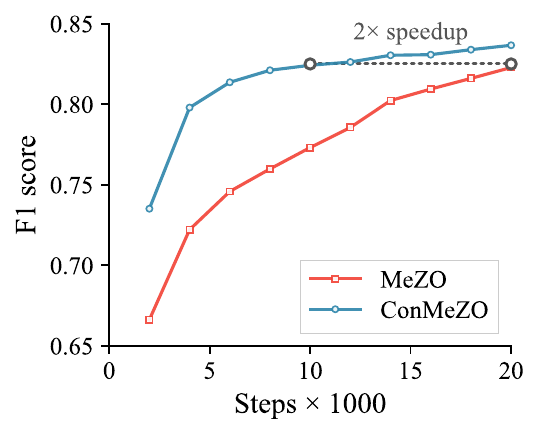}
   \captionof{figure}{ConMeZO requires 2$\times$ fewer training iterations than MeZO to reach the same accuracy when finetuning OPT-1.3B on the SQuAD dataset.}%
  \label{fig:opt_test_acc_squad}
\end{figure}

Finetuning LLMs enables pre-trained models such as LLaMA \citep{touvron2023llama, touvron2023llama2, grattafiori2024llama} and Gemma \citep{team2024gemma, team2024gemma2,team2025gemma} to excel in diverse tasks. By adapting to specific applications, finetuning enhances model capabilities without requiring training from scratch, making state-of-the-art solutions more accessible.
However, traditional finetuning methods face significant challenges due to their high computational and memory demands. These backpropagation-based approaches require substantial GPU resources for storing activation values and computing gradients, often exceeding the budgets when only consumer-grade GPUs are available.

Zeroth-order optimization (ZO) methods, such as those employed by MeZO \citep{malladi2023fine}, offer a promising alternative. By relying only on forward passes to estimate gradients, ZO methods bypass the memory-intensive backward pass, facilitating finetuning in resource-constrained scenarios.
Despite their advantages, ZO methods suffer from high variance in gradient estimates, leading to slower convergence compared to first-order methods. As shown in \citep[Table 15]{malladi2023fine}, while it takes approximately 1K iterations with Adam to finetune a RoBERTa-large model to desirable accuracy, MeZO requires significantly more steps, specifically 100K, to achieve comparable performance. As a result, the overall runtime of MeZO can be significantly longer than that of Adam.

This work aims to address the runtime inefficiency of ZO methods while preserving their memory benefits.
Traditional ZO methods typically rely on random search directions sampled from either a sphere or Gaussian distribution. 
Such random strategies, especially in the high-dimensional regime, result in high variance in gradient estimations and thereby slow convergence. We propose reducing gradient variance by constraining random search directions within a cone centered on a promising search direction, defined by a momentum vector. This strategy improves convergence by narrowing the search space while maintaining the flexibility of ZO optimization.
The proposed approach, coined ConMeZO, significantly reduces iteration counts while retaining the memory efficiency, and matches MeZO in wall-clock runtime. Combining theoretical analysis and empirical validation, we contribute to advancing efficient and accessible finetuning methods for LLMs.

\noindent
\textbf{Contributions:} This work presents a new ZO algorithm, built on an innovative geometrical concept. The algorithm reduces the high variance in gradient estimation by constraining perturbations to a cone centered around a momentum direction. This novel approach balances exploration and exploitation, leading to faster convergence and more reliable ZO optimization. Our contributions can be summarized as:

\begin{enumerate}
\item \textbf{Algorithm design and implementation:} A distinctive cone-sampling strategy inspired by geometrical principles focuses search directions toward areas more likely to yield productive updates. This approach not only reduces noise but also preserves the simplicity of ZO optimization, making it both efficient and theoretically sound. ConMeZO further introduces a vectorized implementation that performs perturbations and updates in fused in-place operations over a flattened parameter buffer, avoiding costly Python loops and random tensor generation. This design yields significant wall-clock speedups without altering the underlying algorithmic logic.
\item \textbf{Theoretical analysis:} Unlike traditional ZO optimizers whose convergence rates suffer from the curse of dimensionality, we show that moderately aligning the momentum to the true gradient can provide up to $O(d)$ speedup over MeZO.
\item \textbf{Improved practical performance:} Experiments on finetuning LLMs demonstrate faster convergence of ConMeZO, especially in early iterations. ConMeZO ultimately achieves up to 2$\times$ speedup over MeZO.

\end{enumerate}

\subsection{Related Work}

\paragraph{Zeroth-order (ZO) optimization.}
The work of \citet{nesterov2017random} marks a foundational step in formally analyzing the convergence rate of zeroth-order methods, such as zeroth-order (stochastic) gradient descent (ZO-SGD) that substitutes gradients in SGD with their zeroth-order estimators.
Building on this foundation, \citet{shamir2017optimal} refine the analysis for nonsmooth convex functions, while \citet{lin2022gradient} extend these insights to nonsmooth nonconvex functions. 
Contributions by \citet{ghadimi2013stochastic} further tackle smooth functions in stochastic settings.
These works have shown that for smooth problems, the squared norm of the gradient converges with a worst-case rate of $O(d/T)$ where $d$ is the number of dimensions \citep{nesterov2017random}. In stark contrast, standard gradient descent has a rate of $O(1/T)$ \citep{nesterov2003introductory}. Further, there exist lower complexity results that prove such dimension dependence is unavoidable \citep{jamieson2012query,wibisono2012finite, duchi2015optimal,golovin2020gradientless,alabdulkareem2021information} unless there are additional structural assumptions such as sparsity or low-rank Hessian \citep{wang2018stochastic,yue2023zeroth}.
More recently, \citet{zhang2025zeroth} prove that zeroth-order methods converge to flat minima for convex and sufficiently smooth functions.

These studies are motivated by the growing interest in zeroth-order methods, driven by practical challenges including the memory limitations imposed by fast differentiation techniques \citep{wang2018stochastic,liu2020primer}.
ZO has been enriched with various enhancements such as 
conditional gradient methods \citep{balasubramanian2018zeroth} and variance reduction techniques \citep{liu2018zeroth, fang2018spider, ji2019improved}. Other notable adaptations include the integration of SignSGD \citep{liu2018signsgd} and applications to minimax optimization \citep{wang2022zeroth}. Beyond algorithmic development, these methods have demonstrated utility across diverse domains, including black-box machine learning \citep{grill2015black, chen2017zoo, chen2019zo}, bandit optimization \citep{flaxman2005online, shamir2017optimal}, reinforcement learning \citep{salimans2017evolution, choromanski2018structured, mania2018simple}, and distributed learning, where they mitigate communication overhead \citep{fang2022communication, zelikman2023just, xu2023federated}.

\paragraph{ZO for LLM finetuning.}

In the realm of ZO optimization for LLMs, various approaches have emerged, emphasizing memory efficiency and computational effectiveness.
MeZO \citep{malladi2023fine} offers a breakthrough by eliminating backpropagation and significantly reducing memory requirements, but it suffers from slower convergence rates and sensitivity to high-dimensional noise.
\citet{zhang2024revisiting} provide a more comprehensive benchmark for evaluating the performance of ZO for LLM finetuning, where they observe that directly combining ZO with momentum methods does not lead to significant performance gain.
\citet{liu2024sparse} introduce sparse MeZO, where only a carefully chosen subset of parameters is updated.
LeZO \citep{wang2024simultaneous} introduces a layer-wise sparse strategy to reduce computational overhead.
\citet{gautam2024variancereduced} integrate variance reduction to ZO optimizers and propose MeZO-SVRG.
\citet{zhao2024second} also use the estimation of second-order information with ZO oracles to improve the performance of MeZO.
Similarly, LOZO \citep{chen2024enhancing} incorporates low-rank gradient estimations, capturing the inherent low-dimensional structure of LLM gradients.
The work of \citet{park2025unraveling} further develops a theoretical framework to characterize effectiveness of structural perturbations, such as sparsity and low rankness, in ZO approaches. 
It is also pointed out in \citet{ma2025revisiting} that effective perturbations in ZO should account for the (estimated) gradient directions, and they propose an approach that requires halving the minibatch data.
DPZero \citep{zhang2024dpzero} extends ZO optimization into the realm of differential privacy, addressing the dual challenges of memory efficiency and data privacy in finetuning LLMs.
Addax \citep{li2024addax} strategically combines first-order and ZO steps to improve overall efficiency.

\paragraph{Notation}

We use $\| \cdot \|$ for the Euclidean norm and $\langle \cdot, \cdot \rangle$ for the standard inner product in $\mathbb R^d$.
Let $\mathbb S^{d-1}= \{ x \in \mathbb R^d \mid \|x \|=1\}$ be the unit sphere in $\mathbb R^d$ and $r\,\mathbb S^{d-1}$ the sphere of radius $r>0$.
A function $f:\mathbb R^d \to \mathbb R$ is $\ell$-smooth iff it is differentiable and $\| \nabla f(x_1)-\nabla f(x_2)\| \leq \ell \|x_1-x_2 \|$, $\forall x_1,x_2 \in \mathbb R^d$, where $\nabla f(x)$ is the gradient at $x$.
The orthogonal complement of a vector $ x \in \mathbb{R}^d $, denoted $ (x)^\perp $, 
is the maximal subspace of $ \mathbb{R}^d $ orthogonal to $ x $, i.e., $ (x)^\perp = \{ v \in \mathbb{R}^d \mid \langle x, v \rangle = 0 \} $.  $\mathcal{N}(0, I_d)$ and $\mathcal{U}(\mathcal{S})$ denote the standard $d$-dimensional Gaussian and the uniform distribution over a set $\mathcal{S}$, respectively.

\section{ZEROTH-ORDER OPTIMIZATION}

We consider solving the following problem with \emph{zeroth-order optimization} (ZO),
\begin{align}
\min_{x \in \mathbb{R}^d} f(x)\,,
\label{eq:zo-opt}
\end{align}
where a direct access to the gradient is unavailable. Instead, we assume access to only a ZO oracle which can compute the objective value at any given point \citep{nesterov2017random}. This setting arises when gradients are challenging or costly to compute \citep{liu2020primer}. The ZO problem is usually solved by applying gradient descent (GD) using a ZO gradient estimator, like the Simultaneous Perturbation Stochastic Approximation (SPSA) \citep{spall1992multivariate}, computed via function value evaluations at perturbed points. We use the following popular stochastic ZO estimator which perturbs the point along randomly sampled directions \citep{nesterov2017random,duchi2015optimal}.
\begin{Definition}
\label{def:spsa}
The stochastic ZO gradient estimate (ZOGE) of $f$ at $x$ using ${z}$ sampled randomly from an isotropic distribution like $\mathcal{N}(0, I_d)$ or $\mathcal{U}(\sqrt{d}\, \mathbb{S}^{d-1})$ and a smoothing parameter $\lambda > 0$ is given by
\begin{align}
g_\lambda(x, z) = \frac{f(x + \lambda {z}) - f(x - \lambda {z})}{2\lambda} z.
\end{align}
\end{Definition}
The ZOGE is highly efficient because it requires only two evaluations of $f(x)$ and avoids explicit gradient computation, thus reducing memory and compute usage for non-trivial $f$. Further, it is known that $\lim_{\lambda \to 0} g_\lambda(x, z)$ is an unbiased estimator of the gradient.
\begin{Lemma}\label{lem:zoge_est}
\citep{zhang2024dpzero}
When $f$ is differentiable and $\lambda$ is sufficiently small, $g_\lambda(x, z) \approx (z^\top \nabla f(x)) z$. Further, the first two moments of this term satisfy
\begin{align}
\mathbb{E}_z[(z^\top \nabla f(x)) z] = \nabla f(x)\; \text{ and }\; \\ \mathbb{E}_z[\|(z^\top \nabla f(x)) z\|^2] \leq 2d \|\nabla f(x)\|^2.
\end{align}
\end{Lemma}
Therefore, despite the benefits mentioned above, this gradient estimator suffers from high $O(d)$ variance, especially in high-dimensional settings of LLM finetuning. It can be shown that the worst case per-step objective function decrease of GD using this estimator is upper-bounded by $O(\|\nabla f(x)\|^2/d)$ \citep{nesterov2017random}. This leads to $O(d)$ slower convergence speed than first-order methods which have $\Omega(\|\nabla f(x)\|^2)$ objective decrease. Addressing this limitation is crucial for making ZO optimization competitive in practical scenarios \citep{malladi2023fine}.
In the next sections, we mitigate this high variance and slow convergence of ZO methods by constraining the search direction $z$ around the true gradient direction estimated via a momentum, while still retaining the memory advantage of ZO methods over first-order approaches.

\section{CONMEZO: ZO WITH CONE SAMPLING} \label{sec:conmezo-design}

To address the slow convergence of ZO methods due to the high variance of the ZOGE (Definition \ref{def:spsa} and Lemma \ref{lem:zoge_est}), we propose a novel cone-based sampling strategy for the search direction $z$ that builds upon the following two components:
\begin{enumerate}
\item \textbf{Promising Search Direction.} A momentum vector $m$ accumulates past gradient estimates, serving as a coarse predictor of future productive search directions; and 
\item \textbf{Cone Restriction.} The perturbation direction $z$ is then constrained to a cone with an apex at the origin, central axis along $m$, and a half-angle $\theta$, reducing the variance of the ZOGE.
\end{enumerate}
The proposed approach can be seamlessly incorporated into ZO methods which use the vanilla ZOGE. 
In \Cref{chap:theoretical}, we show that cone sampling reduces the variance of the gradient estimate by focusing the search in regions more likely to yield productive parameter updates, thereby striking a more effective balance between exploration and exploitation. The next sections formalize the cone-sampling approach and provide an approximate implementation for high-dimensional problems.

\subsection{Momentum-based Search Direction}

\definecolor{myblue}{HTML}{2E86AB}
\definecolor{myred}{HTML}{F24236}
\definecolor{myamber}{HTML}{D4880A}

\begin{figure}[t]
    \centering
    \hspace{1mm}\begin{subfigure}[t]{0.45\linewidth}
        \centering
        \begin{tikzpicture}[scale=0.6]
        
        \tikzset{
            cone/.style={thick, dashed, gray},
            momentum/.style={thick, myblue, ->},
            radius_arrow/.style={<->, thick},
            theta_label/.style={fill=gray!30, inner sep=1pt},
            direction/.style={thick, myred, ->},
            gamma_label/.style={myred, fill=gray!30, inner sep=1pt},
            direction_part/.style={thick, myamber, ->},
        }
        
        \draw[thick] (0,0) circle (3cm);
        
        \draw[radius_arrow] (0,0) -- (3,0) node[midway, above] {$\sqrt{d}$};

        \draw[cone] (0,0) -- (2.12,2.12);
        \draw[cone] (0,0) -- (-2.12,2.12);
        
        \fill[gray!30] (0,0) -- (2.12,2.12) arc[start angle=45,end angle=135,radius=3cm] -- cycle;
        
        \draw[momentum] (0,0) -- (0,3) node[near end, right] {$\sqrt{d}\hat m_t$};
        
        \draw[thick] (0.0,0.0) ++ (90:1.5) arc (90:45:1.5);
        \node[theta_label] at (0.6,1.4) {$\theta$};

        \draw[thick, myred] (0.0, 0.0) ++ (120:1.4) arc (120:90:1.4);
        \node[gamma_label] at (-0.38,1.3) {$\gamma$};
        
        \draw[direction_part] (0,0) -- (-1.5,0) node [midway, below] {$z_t^\perp$};
        
        \draw[direction_part] (-1.5,0) -- (-1.5,2.598) node [near start, left] {$z_t^\parallel$};
        
        \draw[direction] (0,0) -- (-1.5,2.598) node[near end, right] {$z_t$};
        
        \end{tikzpicture}
        \caption{Gray search angle in 2D}
        \label{fig:cone1}
    \end{subfigure}
    \hspace{2mm}
    \begin{subfigure}[t]{0.45\linewidth}
        \centering
        \begin{tikzpicture}[scale=0.95]
    
        \tikzset{
            cone/.style={thick, gray},
            radius_arrow/.style={<->, thick},
            momentum/.style={thick, myblue, ->},
            direction/.style={thick, myred, ->},
        }
    
        \clip (-2,-0.5) rectangle + (4.2,4);
        
        \draw[cone] (0,0) -- (2,2);
        \draw[cone] (0,0) -- (-2,2);
        
        \draw[radius_arrow] (0.1,-0.1) -- (2.1,1.9) node[midway, below right] {$\sqrt{d}$};
        
        \draw[line cap=round, line width=.4mm, myred] (0.0,11.807) ++ (97:-9.3) arc (97:83:-9.3);
        
        \draw[thick, gray] (0.0,0.0) ++ (135:2.828) arc (135:45:2.828);
        
        \draw[thick, dashed, gray] (0.0,-8.8) ++ (100:11) arc (100:80:11);
        
        \draw[momentum] (0,0) -- (0,2.828);
        
        \node [myblue] at (-0.5, 1.5) {$\sqrt{d} \hat m_t$};
        
        \draw[direction] (0,0) -- (1,2.5) node [midway, right] {$z_t$};
        
        \draw[thick, gray] (0.0,11.807) ++ (101.4:-10) arc (101.4:78.6:-10);
        
        \draw[thick] (0.0,0.0) ++ (90:1) arc (90:135:1);
        \node[fill=white, inner sep=1pt] at (-0.4,0.9) {$\theta$};
        
        \draw[thick, myred] (0.0,0.0) ++ (90:1.6) arc (90:68:1.6);
        \node[red, fill=white, inner sep=1pt] at (0.3,1.5) {$\gamma$};
        
        \draw[<->, myamber] (1,2.7) -- (0,2.7) node [above right] {\footnotesize $r(\gamma)$};
        
        \end{tikzpicture}
        \caption{Sampling a $z_t$ in the red ring in the cone.}
        \label{fig:cone3}
    \end{subfigure}
    
    \caption{2D- and 3D-representation of the cone-sampling approach. (a) Sphere with radius $\sqrt{d}$ and (gray) search space cone of half-angle $\theta$ around promising search direction $\hat m_t$. We can set random direction $z_t=z_t^\parallel + z_t^\perp$ with angle $\gamma$ to $\hat m_t$. (b) 3D representation of cone sampling in red area.}
    \label{fig:cone2}
\end{figure}
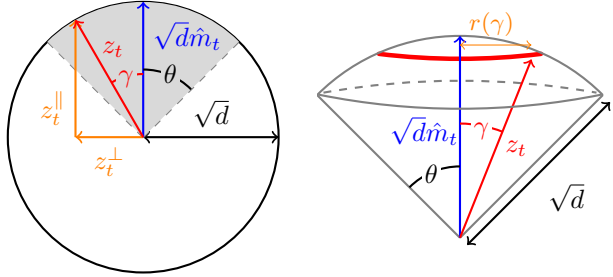

We iteratively construct the promising search direction using a momentum $m_t$ defined as the exponentially moving average of past gradient estimates:
\begin{align}
m_{t+1} \gets \beta \cdot m_t + (1-\beta) \cdot g(x_t, z_t)
\end{align}
where $\beta \in [0,1]$. At each new step, the current gradient estimate $g(x_t, z_t)$ is given a weight $ (1-\beta)$, which serves as a tunable hyperparameter. We adopt a momentum mechanism since it is well known to reduce variance of stochastic GD, particularly in training nonconvex neural networks \citep{tieleman2012lecture,kingma2014adam,cutkosky2019momentum}.

\subsection{Sampling from a Cone}

In this subsection, we discuss how the perturbation direction $z_t$ is sampled uniformly from the intersection of the sphere $\sqrt{d}\, \mathbb{S}^{d-1}$ and the cone with a central axis along the direction ${m}_t$ and half-angle $\theta$. Using the 2D Figure \ref{fig:cone1} as a geometrical reference, we decompose $z_t$ into two additive components:
\begin{enumerate}
    \item $z_t^{\parallel}$, the component of $z_t$ parallel to $m_t$; and,
    \item $z_t^{\perp}$, the component of $z_t$ orthogonal to $ m_t$,
\end{enumerate}
so that $z_t=z_t^{\parallel} + z_t^{\perp}$.
For a fixed angle $\gamma$ between $z_t$ and $m_t$, it follows that 
\begin{align}
z_t^{\parallel}=\sqrt{d} \cos(\gamma) \cdot \hat m_t \;\text{ and }\; z_t^{\perp}=\sqrt{d} \sin(\gamma) \cdot u_t^{\perp}\,,
\end{align}
where $\hat{m}_t = m_t/\|m_t\|$ and $u_t^\perp$ is uniformly sampled from the intersection of the unit sphere $\mathbb{S}^{d-1}$ and the subspace $(m_t)^\perp$ orthogonal to $m_t$. Since $u_t^\perp \perp m_t$, we can directly verify that $\|z_t\|^2 = d$, and thus $z_t \in \sqrt{d} \,\mathbb{S}^{d-1}$.
Next, we discuss how to sample $u_t^\perp$ and introduce randomness in the angle $\gamma$. To this end, we adopt two practical simplifications that we justify in the high-dimensional regime.

\begin{algorithm}[t]
\caption{Cone-based memory-efficient zeroth-order (ConMeZO) optimization algorithm.}\label{alg:a1}
\begin{algorithmic}
\Require Parameters $ x \in \mathbb R^d$, function $f:\mathbb R^d \to \mathbb R$, cone angle $\theta \in [0,\frac \pi  2]$, momentum parameter $\beta \in [0,1]$, iterations $T$, learning rate $\eta$, smoothing parameter $\lambda > 0$.
\For{$t=0, \ldots , T-1$}
\State $u_t \sim \mathcal U(\mathbb S^{d-1})$ \Comment{Sample $u_t$}
\State $[m_0 \gets u_0]_{t=0}$
\State $z_t \gets \sqrt d (\cos(\theta) \cdot m_t / \| m_t \| + \sin (\theta) \cdot u_t )$
\State $ x \gets  x - \eta \cdot g_\lambda (x, z_t)$
\State $m_{t+1} \gets \beta \cdot m_t+ (1-\beta) \cdot g_\lambda (x, z_t)$
\EndFor

\end{algorithmic}
\end{algorithm}

\paragraph{Sampling of orthogonal $u_t^\perp$.}
Instead of sampling $u_t^\perp$ from $\mathbb S^{d-1} \cap (m_t)^\perp$, we sample it uniformly from $\mathbb S ^{d-1}$. This simplification is justified as in high dimensions ($d \gg 1$), a random vector is nearly orthogonal to any fixed direction. We provide a formal result below whose proof is in \cref{app:proof_orth}.
\begin{Proposition}  
    \label{prop:orth}
    The cosine similarity between a randomly sampled vector $u_t \sim \mathcal{U}(\mathbb S^{d-1})$ and a fixed unit vector $\hat{m}_t$ becomes negligible in high dimensions, i.e.,~$\langle \hat{m}_t, u_t \rangle \to 0$ as $d \to \infty$.
\end{Proposition}

\paragraph{Sampling of angle $\gamma$.}
To ensure that $z_t$ is uniformly distributed in the intersection of the sphere and the cone, one must sample $\gamma$ appropriately. However, due to concentration phenomena in high dimensions, most of the probability mass of the distribution of $\gamma$ lies sharply near the edge of the cone.
Thus, in practice, it is suitable to set $\gamma = \theta$. We formalize this intuition below, with the proof given in \Cref{app:proof_angle}.
\begin{Proposition}
\label{prop:angle}
Consider a cone $ \mathcal{C} $ with apex at the origin, central axis aligned with a unit vector $ \hat m_t $, and half-angle $ \theta $. 
If a random vector $ z_t $ is sampled uniformly from $ \mathcal{C} \cap \sqrt{d}\, \mathbb S^{d-1}$,
then the angle $\gamma$ between $z_t$ and $\hat{m}_t$ converges in distribution to a Dirac delta at $\theta$, i.e., $\mathrm{Pr}(\gamma \leq \theta') \to 0$ as $d \to \infty$, $\forall$ $\theta' < \theta$.
\end{Proposition}
The construction of the random direction $z_t$ can be then summarized as follows:
\begin{equation*}
    z_t \leftarrow \sqrt d \Big(\cos(\theta) \cdot \hat m_t + \sin(\theta) \cdot u_t\Big), \; \text{where } u_t \sim \mathcal U(\mathbb S^{d-1}).
\end{equation*}
Note that even though the originally proposed sampling method is conceptually complicated, we have designed a simple and practical approximate implementation for it.
We use the ZOGE (\cref{def:spsa}) with this newly constructed search direction $z_t$ as our gradient estimate.
The resultant ConMeZO method is given in \Cref{alg:a1}.
The impact of hyperparameters $\theta$, $\beta$ and the learning rate $\eta$ will be analyzed theoretically in \Cref{chap:theoretical} and empirically in \Cref{chap:experiments}.

\subsection{Speedup Due to Extra Memory Buffer}

\label{speedup}

Notice that \Cref{alg:a1} maintains an extra optimizer state to store the momentum $m_t$. In our PyTorch implementation, we utilize this extra memory buffer to make ConMeZO faster than MeZO even though the pseudocode of ConMeZO in \Cref{alg:a1} has $2\times$ the number of update steps. To keep the peak GPU VRAM usage small, the MeZO implementation perturbs the iterate one parameter at a time across the neural network and regenerates the random perturbation $z_t$ four times per iteration. 
However, our implementation perturbs the iterate in a single vectorized operation and regenerates $u_t$ only twice because it can temporarily store the perturbation value in the momentum buffer. Note that it is impossible to speed up MeZO similarly unless we increase its memory usage to match that of ConMeZO.  \Cref{tab:run_time} confirms that our implementation of ConMeZO is faster than that of MeZO. More details and sample code are provided in \Cref{sec:speedups}. 

\subsection{Momentum Warm-up}

Since the initial perturbation directions are random and not necessarily informative, they can bias the trajectory of the ConMeZO optimizer for many future steps. To address this, we gradually increase the momentum parameter $\beta$ during the early phase of training. This allows the momentum to be shaped by multiple directions before it becomes dominant, leading to more stable early iterations and more reliable progress later on.
In practice, we use the following warm-up schedule for a training run of 20K steps, which we find to work well:
\[
\beta_t =
\begin{cases}
0.1, & 0 \le t \le 200, \\[6pt]
\beta_{\text{final}} - \dfrac{\beta_{\text{final}} - 0.1}{\bigl(1 + 8 \cdot \left(\frac{t-200}{1800}\right)^{1.8} \bigr)^{3}}, & 200 < t \le 2000, \\[10pt]
\beta_{\text{final}}, & t > 2000,
\end{cases}
\]

For shorter training runs of 10K steps, we simply halve the interval lengths (i.e., 0--100, 100--1000, and beyond).
This schedule smoothly transitions from a small initial value to the final momentum parameter, stabilizing early optimization without additional overhead.
The schedule is designed to have three phases: (i) a short flat start to avoid immediate momentum bias, (ii) a smooth ease-in ramp to gradually build momentum, and (iii) a quick saturation to the final value. This shape is chosen empirically and gives stable early optimization and consistent performance across tasks.
Detailed intuition, schedule visualization (\Cref{fig:warmup_schedule}), and ablation results (\Cref{tab:warmup_ablation}) are provided in \Cref{app:warmup_details}.
\section{THEORETICAL ANALYSIS}
\label{chap:theoretical}

In this section, we formally analyze ConMeZO (\Cref{alg:a1}) for optimization problems with $\ell$-smooth and potentially nonconvex objectives. Our analysis shows that at each iteration, ConMeZO can decrease the objective value faster than the standard ZO method, MeZO \citep{malladi2023fine}, provided that the momentum is well aligned with the true gradient. Furthermore, the convergence rate of ConMeZO is no worse than that of MeZO.
We begin our analysis by characterizing the first and second moments of the cone-based gradient estimator used in ConMeZO (\Cref{alg:a1}). For ease of exposition, we assume that the ``smoothing'' parameter $\lambda$ of the ZOGE (\cref{def:spsa}) estimator is infinitesimally small, i.e., $\lambda \to 0$.

\begin{Lemma}
\label{lem:first_moment}
Let $a_t = \nabla f(x_t)$, $\rho_t$ be the angle between $a_t$ and ${m}_t$, and $z_t = \sqrt{d} (\cos (\theta) \cdot \hat{m}_t + \sin ( \theta ) \cdot u_t)$, where $u_t \sim \mathcal{U}(\mathbb S^{d-1})$. When $\lambda \to 0$, the ZOGE becomes $(z_t^\top a_t) z_t$, and its first and second moments satisfy:
\begin{align*}
\bE_{u_t} [ &(z_t ^\top a_t) z_t ] = d\cos^2 \theta \cdot (\hat m_t^\top a_t) \hat m_t + \sin^2 \theta \cdot a_t\,, \text{ and } \\
\bE_{u_t} \big[ &\| (z_t^\top a_t)z_t \|^2 \big] \\
&\leq d \|a_t \|^2 \big((d+4) \cos^2 \theta \cos^2 \rho_t + \sin^2 \theta \big)\,.
\end{align*}
\end{Lemma}
A proof can be found in \Cref{app:proof_first_moment}. 
Comparing these results  with the moments of the vanilla ZOGE  in \Cref{lem:zoge_est}, we see that the cone-sampled gradient estimator is biased toward the momentum direction $\hat m_t$, and the second moment has an extra $O(\cos^2 \rho_t)$ term. The bias vanishes and the second moment becomes similar to that in \Cref{lem:zoge_est} when $\hat m_t$ aligns with the true gradient $a_t$, i.e., $\rho_t \approx 0$. The parameter $\theta$ modulates the tradeoff between the momentum-aligned component and the true gradient component.

\subsection{Convergence Guarantee}

Next, we characterize the expected improvement per iteration and the global convergence of ConMeZO. Using \Cref{lem:first_moment}, we get the following ``descent lemma'' whose proof is in \Cref{app:proof_improvement_per_it}.

\begin{Theorem}[Descent Lemma]
\label{thm:improvement_per_it}
Assume $f$ is $\ell$-smooth and let $\lambda\to0$. The expected value of \(f(x_{t+1})\) satisfies
\begin{align*}
& \bE_{u_t}[f(x_{t+1})] \leq f(x_t) \\
& - \eta \left(d \cos^2 (\theta) \cos^2 (\rho_t) + \sin^2 (\theta) \right) \|a_t\|^2 \\
& + \frac{\eta^2 \ell}{2} d \left((d+4) \cos^2 (\theta) \cos^2 (\rho_t) + \sin^2 (\theta) \right) \|a_t\|^2.
\end{align*}
Further, with the choice of $\eta = (d \cos^2 (\theta) \cos^2 (\rho_t) + \sin^2 (\theta))/(\ell d \left((d+4) \cos^2 (\theta) \cos^2 (\rho_t) + \sin^2 (\theta)\right))$, we can get that
\begin{align*}
&\bE_{u_t}[f(x_{t+1})] - f(x_t) \\
&\leq - \frac{\l( d \cos^2(\theta) \cos^2(\rho_t) + \sin^2(\theta) \r)^2}{2 \ell d \l( (d+4)\cos^2(\theta)\cos^2(\rho_t) + \sin^2(\theta)\r)} \|a_t\|^2 \\
&\approx -  \frac{ (d \cos^2(\theta) \cos^2(\rho_t) + \sin^2(\theta))}{2 \ell d} \|a_t\|^2.
\end{align*}
\end{Theorem}

The above theorem shows that ConMeZO achieves a per-step decrease on the order of $O(\|a_t\|^2)$, provided that the momentum $\hat{m}_t$ is effectively aligned with the gradient $a_t$.
Specifically, when the alignment is strong (i.e., $\cos^2(\rho_t)$ is large) and $\sin(\theta)$ is small, ConMeZO can significantly outperform standard MeZO, which is restricted to a $O(\|a_t\|^2/d)$ descent \citep{nesterov2017random}. Thus, the progress of ConMeZO aligns directly with how well the momentum tracks the gradient direction. We empirically verify this alignment between the momentum vector and the true gradient in \Cref{additional_results} (\Cref{fig:cos}).

\paragraph{Trade-off in $\theta$.}
The cone angle $\theta$ balances the exploitation of the promising momentum direction and the exploration of alternative directions through random sampling. At step $t$, the maximally exploitative choice for $\theta$ is
\begin{align}
\theta^\star =
\begin{cases}
0, & \text{if } \cos^2(\rho_t) > (d+4)/d^2, \\
\frac \pi 2, & \text{otherwise.}
\end{cases}
\end{align}
Since the expected squared cosine similarity between two random vectors is $1/d \approx (d+4)/d^2$, this result indicates that momentum should be exploited precisely when it provides a better-than-random direction, i.e., when $\cos^2(\rho_t) > 1/d$.
However, this binary prescription is extreme and does not account for the global convergence as $\theta$ affects the quality of future momentum vectors. A small $\theta$ reduces exploration, while a large $\theta$ sacrifices exploitation. In practice, it is better to use a balanced choice of $\theta$, which maintains some directional bias while still allowing sufficient exploration. Empirical observations in Section \ref{additional_results} confirm this trade-off.
We now establish an overall convergence guarantee of our algorithm over $T$ iterations.
\begin{Corollary}
\label{cor:convergence}
Assume $f$ is $\ell$-smooth and bounded from below, i.e.,~$f^\star = \min_{x \in \mathbb R^d} f(x) > -\infty$ and let $\lambda \to 0$. There exists a hyperparameter setting for ConMeZO such that after $T \geq 1$ steps, the average expected squared gradient norm over $T$ iterations satisfies
\begin{align}
    \frac{1}{T} \sum_{t=0}^{T-1} \mathbb{E} \left[ \|\nabla f(x_t)\|^2 \right] \leq \frac{2 \ell d (f(x_0) - f^\star)}{T}.
\end{align}
\end{Corollary}
A proof directly follows from \Cref{thm:improvement_per_it} by setting $\theta=\pi/2$ and telescoping across $T$ iterations. 
This establishes that ConMeZO matches the worst-case convergence rate of MeZO, while Theorem \ref{thm:improvement_per_it} shows it can be significantly faster whenever momentum aligns better than chance with the gradient ($\cos^2(\rho_t)\gg 1/d$).
\section{EXPERIMENTAL RESULTS}
\label{chap:experiments}

In this section, we study the empirical performance and efficiency of ConMeZO on a synthetic problem and LLM finetuning tasks
with RoBERTa \citep{liu2019roberta} and OPT \citep{zhang2022opt} backbones. Our code can be found at \url{https://github.com/LejsDeen/ConMeZO}.

\begin{figure}[]
  \centering
      \includegraphics[width=0.7\linewidth]{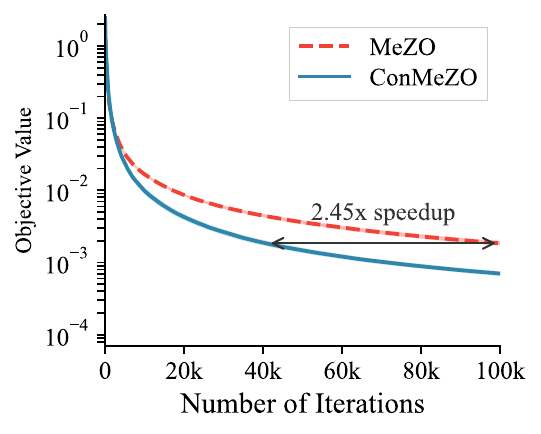}
   \captionof{figure}{\textbf{Synthetic Optimization Problem:} ConMeZO achieves a 2.45$\times$ speedup over MeZO on the synthetic quadratic problem.}%
  \label{fig:general_quadratic}
\end{figure}

\begin{table}[]
\centering
\caption{\textbf{RoBERTa-Large:} ConMeZO achieves better test metrics (\%) than MeZO, and MeZO+Momentum (Mom.) when finetuning RoBERTa-large.}
\begin{tabular}{lcccc}
\toprule
& & \multicolumn{3}{c}{ZO Methods:} \\
\cmidrule(lr){3-5}
Task   & AdamW & MeZO & Mom. & ConMeZO \\
\midrule
SST-2  & 93.1 & 92.8 & 92.2 & \textbf{93.5} \\
SST-5  & 56.6 & 49.3 & \textbf{51.9} & 48.9 \\
SNLI   & 86.4 & 81.0 & 80.9 & \textbf{81.9} \\
MNLI   & 81.4 & 69.7 & 70.7 & \textbf{73.2} \\
RTE    & 83.6 & 73.9 & 74.0 & \textbf{75.1} \\
TREC   & 95.9 & 88.4 & 89.2 & \textbf{90.0} \\
\midrule
Average & 82.8 & 75.8 & 76.5 & \textbf{77.1} \\
\bottomrule
\end{tabular}
\label{tab:performance_10k}
\end{table}

\begin{table*}[ht]
\centering
\caption{\textbf{OPT-1.3B and OPT-13B}: ConMeZO achieves better average test accuracy / F1 score (\%) than MeZO when averaged over 3 seeds. OOM means out of memory.}
\setlength{\tabcolsep}{4pt}
\begin{tabular}{llccccccccc}
\toprule
Model & Method & SQuAD & SST2 & WIC & BoolQ & DROP & ReCoRD & RTE & MultiRC & Avg. \\
\midrule
\multirow{2}{*}{OPT-1.3B} & MeZO     & 72.76 & 88.49 & 56.53 & 63.50 & 25.90 & 70.67 & \textbf{56.92} & \textbf{55.90} & 61.33 \\
& ConMeZO  & \textbf{75.34} & \textbf{90.56} & \textbf{58.15} & \textbf{64.20} & \textbf{26.53} & 70.67 & 55.48 & 53.50 & \textbf{61.80} \\
\midrule
\multirow{2}{*}{OPT-13B}  & MeZO     & 82.28 & 91.25 & \textbf{58.93} & 67.73 & OOM & \textbf{81.17} & 63.66 & 57.53 & 71.79 \\
& ConMeZO  & \textbf{83.66} & \textbf{92.39} & 58.31 & \textbf{69.33} & OOM & 80.87 & \textbf{64.50} & \textbf{58.30} & \textbf{72.48} \\
\bottomrule
\end{tabular}
\label{tbl:opt_combined}
\end{table*}

\subsection{Synthetic Optimization Problem}
\label{sec:synthetic_experiments}

First, we compare MeZO and ConMeZO on a synthetic strongly convex quadratic problem in $d=1000$ dimensions with a condition number of value $d$. We tune the hyperparameters of both the methods over a simple grid with the final average objective value over $5$ trials serving as selection criteria. In \Cref{fig:general_quadratic}, we plot the mean objective value (and standard error) of the best hyperparameter settings for each method. We observe that even in this simple synthetic problem, ConMeZO achieves a 2.45$\times$ speedup over MeZO. See Appendix~\ref{apdx.sec.synthetic} for more details.

\subsection{Finetuning of RoBERTa-large}
\label{sec:roberta_experiments}

We finetune RoBERTa-large \citep{liu2019roberta}, a language model with 355 million parameters. Following \cite{malladi2023fine},
the finetuning process tackles a few-shot setting on a suite of six standard NLP tasks from the GLUE benchmark \citep{wang2018glue}. We compare ConMeZO with first-order AdamW, MeZO, and the MeZO+Momentum baseline. MeZO+Momentum is a novel baseline that we design, and it maintains a momentum $m_t$ similar to ConMeZO, but instead of using it to bias the ZO perturbation $z_t$, its momentum replaces $g(x_t, z_t)$ as the iterate update direction. We set the smoothing parameter as $\lambda = 10^{-3}$ for all ZO methods and report the test metrics after tuning the hyperparameters for 10K iterations. See Appendix~\ref{apdx.sec.roberta} for more details.
In \Cref{tab:performance_10k}, ConMeZO achieves the best average performance across all tasks except SST-5, where MeZO+Momentum performs best. While MeZO+Momentum outperforms vanilla MeZO on average, it does not consistently match ConMeZO, suggesting that naively incorporating momentum into ZO methods is suboptimal. 
We report standard errors (\Cref{tab:performance_10k_std}) along with intermediate metrics (\Cref{tab:acc_comparison_dev}) and a comparison to first-order SGD (\Cref{tab:performance_10k_sgd}) in \Cref{apdx.sec.roberta}. Additional analyses, including test metric curves (\Cref{fig:test_acc_10k}) and an ablation study of parameters $\beta$ and $\theta$ (\Cref{fig:heatmaps}), can be found in \Cref{additional_results}.

\begin{figure}[tb]
  \centering
  \includegraphics[width=\linewidth]{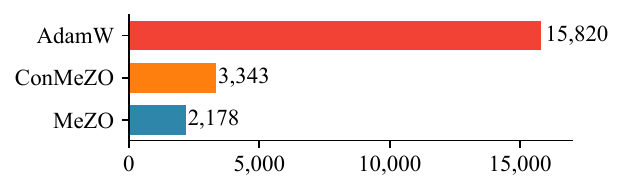}
  \vspace{-1em}
  \includegraphics[width=\linewidth]{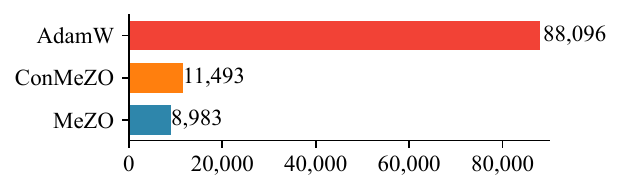}
  \caption{Peak GPU memory usage (MiB) increase of ConMeZO over MeZO is negligible when compared to the memory usage of first-order methods like AdamW: \textbf{Top:} RoBERTa-Large on SST2 (batch size 64). \textbf{Bottom:} OPT-1.3B on BoolQ (batch size 16).}
  \label{fig:gpu_mem}
\end{figure}

\subsection{Finetuning of OPT models}
\label{sec:opt_experiments}

Here, we finetune OPT‐1.3B and OPT-13B \citep{zhang2022opt} on eight standard benchmarks using ConMeZO and MeZO. We omit AdamW due to GPU VRAM limitations, and MeZO+Momentum since it did not achieve the best performance on RoBERTa.
We set the smoothing parameter as $\lambda = 10^{-3}$ for both methods and report the test metrics after tuning the hyperparameters for 20K iterations. See Appendix~\ref{apdx.sec.opt} for more details.
In \Cref{tbl:opt_combined}, we see that ConMeZO delivers the highest final accuracy/F1 on almost every task and achieves the best average across the tasks. The advantages of ConMeZO we observed in RoBERTa carry over seamlessly to the larger OPT-1.3B and OPT-13B models.
Note that we omit the DROP dataset for OPT-13B because we obtained an out-of-memory error when running both MeZO and ConMeZO on our GPU.
We also plot learning curves of ConMeZO and MeZO when finetuning OPT-1.3B on the SQuAD dataset in \Cref{fig:opt_test_acc_squad}.
The steep slope highlights ConMeZO's accelerated convergence: it reaches MeZO's 20K step performance in less than 10K steps, which yields a 2$\times$ speedup.
Standard errors and full task breakdowns are reported in \Cref{apdx.sec.opt} (\Cref{tbl:opt_1.3b_std} for OPT-1.3B and \Cref{tbl:opt_13b_std} for OPT-13B).

\subsection{Compute and Memory Efficiency}

Compared to MeZO, ConMeZO achieves a slightly lower runtime per iteration while incurring a moderate increase in memory consumption.
For instance, on RoBERTa-large with SST-2, ConMeZO requires \texttt{3343 MiB} versus MeZO’s \texttt{2178 MiB}, a relative increase of $+54\%$. Importantly, this overhead becomes much less pronounced at larger context lengths, which is consistent with current trends in LLM training, where the relative gap shrinks significantly. As shown in \Cref{fig:gpu_mem} and \Cref{tab:mem_combined}, \Cref{sec:speedups}, the increase is only around $+10\%$ on OPT-1.3B with DROP and $+28\%$ with BoolQ, while still staying within the same order of magnitude as MeZO. In contrast, first-order optimizers such as AdamW demand dramatically more resources (e.g., \texttt{15820 MiB} for finetuning RoBERTa-Large on SST-2; see \Cref{fig:gpu_mem}).
The runtime comparison in \Cref{tab:run_time} further highlights that ConMeZO not only avoids extra cost but is in fact faster per iteration, despite its seemingly higher algorithmic complexity. On average, ConMeZO improves iteration speed by $3.6\%$ on RoBERTa-Large and $7.9\%$ on OPT-1.3B.
This speedup comes from the extra memory buffer as discussed in \Cref{sec:conmezo-design}.

\begin{table*}[tp]
\centering
\caption{ConMeZO achieves better average wall-clock time (s) per step across all the tasks than MeZO when finetuning RoBERTa-large and OPT-1.3B.}
\label{tab:run_time_combined}
\setlength{\tabcolsep}{4pt}
\begin{tabular}{lcccccccccccc}
\toprule
 & \multicolumn{7}{c}{RoBERTa-Large} & \multicolumn{5}{c}{OPT-1.3B} \\
\cmidrule(lr){2-8} \cmidrule(lr){9-13}
Method & SST2 & SST5 & SNLI & MNLI & RTE & TREC & Avg. & SST2 & BoolQ & DROP & SQuAD & Avg. \\
\midrule
MeZO     & 0.193 & 0.202 & 0.198 & 0.335 & 0.548 & 0.146 & 0.270 
& 0.096  & 0.270 & 0.426 & 0.194 & 0.246 \\
ConMeZO  & \textbf{0.180} & \textbf{0.195} & \textbf{0.189} & \textbf{0.325} & \textbf{0.538} & \textbf{0.144} & \textbf{0.262}
& \textbf{0.083} & \textbf{0.257} & \textbf{0.413} & \textbf{0.181} & \textbf{0.233} \\
\midrule
\% Speedup  & 6.95\%  & 3.59\%  & 4.89\%  & 3.15\%  & 1.89\%  & 1.10\%  & 3.60\% 
& 15.73\% & 5.10\%  & 3.33\%  & 7.29\%  & 7.86\% \\
\bottomrule
\end{tabular}
\label{tab:run_time}
\end{table*}
\section{COMPARISON WITH RECENT ZO METHODS}
\label{sec:comparison_recent_zo}

We compare ConMeZO with recent ZO finetuning methods to clarify the relationship and demonstrate empirical performance. These methods target different aspects of ZO optimization and are largely orthogonal to ConMeZO's cone-guided directional sampling.

\subsection{HiZOO: Second-Order Information}

HiZOO \citep{zhao2024second} augments MeZO with second-order curvature information via three function evaluations per step, estimating local Hessian structure to refine update directions. In principle, ConMeZO's cone-guided sampling could be combined with HiZOO's curvature-informed estimator.

\textbf{Test accuracy.}
For RoBERTa-Large, we evaluated both methods for 10k steps (yielding $\sim$2$\times$ longer wall-clock time for HiZOO), while for OPT-1.3B we used equal wall-clock budgets (HiZOO 10K steps, ConMeZO 20K steps). For HiZOO, we use settings recommended in their paper; ConMeZO uses our default configuration (\Cref{apdx.sec.roberta,apdx.sec.opt}).
See results in \Cref{tab:hizoo_accuracy}.

\begin{table}[h]
\centering
\small
\caption{ConMeZO maintains a clear advantage across the majority of tasks and both models. For HiZOO, we report the best performance across learning rates $\{10^{-5}, 10^{-6}, 10^{-7}\}$ selected individually for each task.}
\setlength{\tabcolsep}{4pt} 
\begin{tabular}{l ccccc c}
\toprule
\multicolumn{3}{c}{RoBERTa-Large (5 seeds)} & \multicolumn{3}{c}{OPT-1.3B (3 seeds)} & \\
\cmidrule(lr){1-3} \cmidrule(lr){4-6}
Method & SST-2 & RTE & SST-2 & BoolQ & WiC & Avg. \\
\midrule
HiZOO & 83.4 & 54.2 & 88.5 & 63.0 & \textbf{59.0} & 69.6 \\
ConMeZO & \textbf{93.5} & \textbf{75.1} & \textbf{90.6} & \textbf{64.2} & 58.2 & \textbf{76.3} \\
\bottomrule
\end{tabular}
\label{tab:hizoo_accuracy}
\end{table}

HiZOO performs competitively on OPT-1.3B but exhibits sensitivity to hyperparameters on RoBERTa-Large. ConMeZO maintains robust performance across tasks with fixed hyperparameters.

\textbf{Wall-clock time and VRAM.} We evaluate both methods on OPT-1.3B with identical hardware and batch size for 2K steps. ConMeZO is 2--2.25$\times$ faster than HiZOO across tasks ($\sim$4 vs $\sim$9 minutes on SST-2, $\sim$18 vs $\sim$36 minutes on BoolQ), due to HiZOO's higher per-step overhead. VRAM usage is comparable: ConMeZO peaks at \texttt{11493} MiB while HiZOO uses \texttt{10681} MiB on OPT-1.3B/SST-2.

\subsection{LOZO: Low-Rank Perturbations}

LOZO \citep{chen2024enhancing} incorporates low-rank structure into gradient estimations, restricting updates to a low-dimensional adapter subspace. In principle, one could apply ConMeZO's cone-guided ZO estimator within LOZO's low-rank adapter space.

We compare LOZO, its momentum variant LOZO-M, and ConMeZO on RoBERTa-Large under equal wall-clock time. LOZO has $\sim$20\% higher per-step runtime than ConMeZO (0.022s vs 0.018s per step on SST-2). Following the authors' recommendations, we sweep learning rates $\{1\text{e-}6, 1\text{e-}7\}$, rank $\{1, 2\}$, and update interval $\nu \in \{50, 100\}$. ConMeZO uses default settings. We provide a comparison in \Cref{tab:lozo_comparison}.

\begin{table}[h]
\centering
\setlength{\tabcolsep}{2pt}
\small
\caption{Accuracy comparison on RoBERTa-Large. Under equal wall-clock time, ConMeZO achieves superior average performance compared to LOZO \& LOZO-M.}
\begin{tabular}{lccccccc}
\toprule
Method & SST2 & SST5 & SNLI & MNLI & RTE & TREC & Avg \\
\midrule
LOZO & 92.3 & 49.4 & \textbf{82.5} & 70.5 & \textbf{78.7} & 89.4 & 77.1  \\
LOZO-M & 92.8 & \textbf{50.4} & 81.3 & 69.5 & 75.1 & 89.8 & 76.5 \\
ConMeZO & \textbf{93.2} & 50.0 & 82.0 & \textbf{73.3} & 76.2 & \textbf{90.4} & \textbf{77.5}  \\
\bottomrule
\end{tabular}
\label{tab:lozo_comparison}
\end{table}

\subsection{MeZO-SVRG: Variance Reduction}

MeZO-SVRG \citep{gautam2024variancereduced} reduces \emph{data-induced} variance by periodically computing large-batch reference gradients. This targets a different variance source than ConMeZO's \emph{directional} variance reduction, making the methods conceptually orthogonal and potentially combinable.

Importantly, the experiments in \citep{gautam2024variancereduced} are conducted in the non-prompted fine-tuning setting, whereas our work focuses on the prompt-conditioned regime. To ensure a direct comparison under our target regime, we evaluate MeZO-SVRG in the same prompt-conditioned setting used throughout this paper.

We evaluate on RoBERTa-Large in the prompt-conditioned setting (3 seeds) in Table \ref{tab:mezo_svrg}.

\begin{table}[h]
\centering
\small
\caption{Accuracy (\%). MeZO-SVRG was run for 24K steps on both tasks,
ConMeZO for 10K (SST-2) and 20K (MNLI) steps. ConMeZO matches or exceeds MeZO-SVRG's accuracy with fewer training steps.}
\begin{tabular}{lcc}
\toprule
Method & SST-2 & MNLI \\
\midrule
MeZO-SVRG & 92 & 75 \\
ConMeZO & \textbf{93.5} & \textbf{76.4} \\
\bottomrule
\end{tabular}
\label{tab:mezo_svrg}
\end{table}

\textbf{Wall-clock time and VRAM.} MeZO-SVRG computes a full-batch ZO gradient every other iteration, requiring $\sim$16 minutes per 100 steps versus ConMeZO's $\sim$1 minute. VRAM usage is similar: \texttt{3719} MiB for MeZO-SVRG vs \texttt{3343} MiB for ConMeZO (RoBERTa-Large, SST-2).

\subsection{ZO-AdaMM: Adaptive Moments}

\begin{table}[h]
\centering
\small
\setlength{\tabcolsep}{10pt}
\caption{Accuracy (\%) comparison on SST-2. ConMeZO consistently outperforms ZO-AdaMM across model architectures. ZO-AdaMM results are as reported by \citet{zhang2024revisiting} and always use 20K steps; ConMeZO results use 10K steps for RoBERTa-Large and 20K steps for OPT-1.3B.}
\begin{tabular}{lcc}
\toprule
Method & RoBERTa-Large & OPT-1.3B \\
\midrule
ZO-AdaMM & 89.8 & 84.4 \\
ConMeZO & \textbf{93.5} & \textbf{90.6} \\
\bottomrule
\end{tabular}
\label{tab:zo_adamm_comp}
\end{table}

ZO-AdaMM stores a second-moment buffer, increasing memory usage beyond ConMeZO. Conceptually, one could apply ZO-AdaMM's adaptive scaling on top of ConMeZO's cone-guided estimates, potentially improving robustness at the cost of additional memory.
\section{CONCLUSION}

This work explores the challenges and opportunities in finetuning LLMs using ZO optimization. By introducing a novel cone-sampling strategy, we propose ConMeZO that mitigates the high variance of traditional random-direction estimators and leverages momentum to guide updates more effectively. Empirical evaluations on RoBERTa-large and larger OPT-1.3B and OPT-13B models
show that our method consistently outperforms state-of-the-art ZO optimizers like MeZO, achieving up to a 2$\times$ speedup in early convergence and absolute accuracy gains across benchmarks.
While ConMeZO incurs a modest memory overhead compared to MeZO, this trade-off is practical for settings where gradient access is unavailable or expensive.

ConMeZO's directional sampling strategy is largely orthogonal to recent ZO methods which target complementary aspects. This suggests potential for hybrid approaches that combine ConMeZO's cone-guided sampling with these techniques. Future work can explore such combinations, as well as lightweight, self-adaptive mechanisms for adjusting the cone angle $\theta$ and momentum $\beta$ dynamically during optimization. Finally, establishing theoretical guarantees that ConMeZO achieves strictly better convergence rates than MeZO under realistic assumptions remains a valuable direction.

\clearpage

\subsection*{Acknowledgements}
L.Z. gratefully acknowledges funding by the Max Planck ETH Center for Learning Systems (CLS). B.L. is supported by SNSF Project Funding No. 200021-207343.
This work does not relate to the current position of K.T. at Amazon.

\bibliographystyle{apalike}
\bibliography{refs}

@inproceedings{alabdulkareem2021information,
  title={Information-theoretic lower bounds for zero-order stochastic gradient estimation},
  author={Alabdulkareem, Abdulrahman and Honorio, Jean},
  booktitle={IEEE International Symposium on Information Theory (ISIT)},
  pages={2316--2321},
  year={2021},
  organization={IEEE}
}

@article{balasubramanian2018zeroth,
  title={Zeroth-order (non)-convex stochastic optimization via conditional gradient and gradient updates},
  author={Balasubramanian, Krishnakumar and Ghadimi, Saeed},
  journal={Advances in Neural Information Processing Systems},
  volume={31},
  year={2018}
}

@misc{bentivogli2009fifth,
  title={The Fifth {PASCAL} Recognizing Textual Entailment Challenge},
  author={Bentivogli, Luisa and Clark, Peter and Dagan, Ido and Giampiccolo, Danilo},
  year={2009},
}

@inproceedings{bowman2015large,
  title={A large annotated corpus for learning natural language inference},
  author={Bowman, Samuel R and Angeli, Gabor and Potts, Christopher and Manning, Christopher D},
  booktitle={Proceedings of the Conference on Empirical Methods in Natural Language Processing},
  pages={632--642},
  year={2015}
}

@inproceedings{chen2017zoo,
  title={{ZOO}: Zeroth order optimization based black-box attacks to deep neural networks without training substitute models},
  author={Chen, Pin-Yu and Zhang, Huan and Sharma, Yash and Yi, Jinfeng and Hsieh, Cho-Jui},
  booktitle={Proceedings of the ACM Workshop on Artificial Intelligence and Security},
  pages={15--26},
  year={2017}
}

@article{chen2019zo,
  title={{ZO-AdaMM}: Zeroth-order adaptive momentum method for black-box optimization},
  author={Chen, Xiangyi and Liu, Sijia and Xu, Kaidi and Li, Xingguo and Lin, Xue and Hong, Mingyi and Cox, David},
  journal={Advances in Neural Information Processing Systems},
  volume={32},
  year={2019}
}

@inproceedings{chen2024enhancing,
  title={Enhancing Zeroth-order Fine-tuning for Language Models with Low-rank Structures},
  author={Yiming Chen and Yuan Zhang and Liyuan Cao and Kun Yuan and Zaiwen Wen},
  booktitle={International Conference on Learning Representations},
  year={2025}
}

@inproceedings{choromanski2018structured,
  title={Structured evolution with compact architectures for scalable policy optimization},
  author={Choromanski, Krzysztof and Rowland, Mark and Sindhwani, Vikas and Turner, Richard and Weller, Adrian},
  booktitle={International Conference on Machine Learning},
  pages={970--978},
  year={2018},
  organization={PMLR}
}

@inproceedings{clark2019boolq,
  title={{BoolQ}: Exploring the Surprising Difficulty of Natural Yes/No Questions},
  author={Clark, Christopher and Lee, Kenton and Chang, Ming-Wei and Kwiatkowski, Tom and Collins, Michael and Toutanova, Kristina},
  booktitle={Proceedings of the Conference of the North American Chapter of the Association for Computational Linguistics},
  pages={2924--2936},
  year={2019}
}

@article{cutkosky2019momentum,
  title={Momentum-based variance reduction in non-convex {SGD}},
  author={Cutkosky, Ashok and Orabona, Francesco},
  journal={Advances in Neural Information Processing Systems},
  volume={32},
  year={2019}
}

@misc{dagan2005pascal,
  title={The {PASCAL} recognising textual entailment challenge},
  author={Dagan, Ido and Glickman, Oren and Magnini, Bernardo},
  year={2005},
}

@inproceedings{dua2019drop,
  title={{DROP}: A Reading Comprehension Benchmark Requiring Discrete Reasoning Over Paragraphs},
  author={Dua, Dheeru and Wang, Yizhong and Dasigi, Pradeep and Stanovsky, Gabriel and Singh, Sameer and Gardner, Matt},
  booktitle={Proceedings of the Conference of the North American Chapter of the Association for Computational Linguistics},
  pages={2368--2378},
  year={2019}
}

@article{duchi2015optimal,
  title={Optimal rates for zero-order convex optimization: The power of two function evaluations},
  author={Duchi, John C and Jordan, Michael I and Wainwright, Martin J and Wibisono, Andre},
  journal={IEEE Transactions on Information Theory},
  volume={61},
  number={5},
  pages={2788--2806},
  year={2015},
  publisher={IEEE}
}

@article{fang2018spider,
  title={{SPIDER}: Near-optimal non-convex optimization via stochastic path-integrated differential estimator},
  author={Fang, Cong and Li, Chris Junchi and Lin, Zhouchen and Zhang, Tong},
  journal={Advances in Neural Information Processing Systems},
  volume={31},
  year={2018}
}

@article{fang2022communication,
  title={Communication-efficient stochastic zeroth-order optimization for federated learning},
  author={Fang, Wenzhi and Yu, Ziyi and Jiang, Yuning and Shi, Yuanming and Jones, Colin N and Zhou, Yong},
  journal={IEEE Transactions on Signal Processing},
  volume={70},
  pages={5058--5073},
  year={2022},
  publisher={IEEE}
}

@inproceedings{flaxman2005online,
  title={Online convex optimization in the bandit setting: {G}radient descent without a gradient},
  author={Flaxman, Abraham D and Kalai, Adam Tauman and McMahan, H Brendan},
  booktitle={Proceedings of the ACM-SIAM Symposium on Discrete Algorithms},
  pages={385--394},
  year={2005}
}

@inproceedings{gautam2024variancereduced,
  title={Variance-reduced Zeroth-Order Methods for Fine-Tuning Language Models},
  author={Gautam, Tanmay and Park, Youngsuk and Zhou, Hao and Raman, Parameswaran and Ha, Wooseok},
  booktitle={International Conference on Machine Learning},
  pages={15180--15208},
  year={2024},
  organization={PMLR}
}

@article{ghadimi2013stochastic,
  title={Stochastic first-and zeroth-order methods for nonconvex stochastic programming},
  author={Ghadimi, Saeed and Lan, Guanghui},
  journal={SIAM Journal on Optimization},
  volume={23},
  number={4},
  pages={2341--2368},
  year={2013},
  publisher={SIAM}
}

@misc{giampiccolo2007third,
  title={The third {PASCAL} recognizing textual entailment challenge},
  author={Giampiccolo, Danilo and Magnini, Bernardo and Dagan, Ido and Dolan, William B},
  year={2007}
}

@inproceedings{golovin2020gradientless,
  title={Gradientless Descent: High-Dimensional Zeroth-Order Optimization},
  author={Daniel Golovin and John Karro and Greg Kochanski and Chansoo Lee and Xingyou Song and Qiuyi Zhang},
  booktitle={International Conference on Learning Representations},
  year={2020}
}

@article{grattafiori2024llama,
  title={The {LLaMA} 3 herd of models},
  author={Grattafiori, Aaron and Dubey, Abhimanyu and Jauhri, Abhinav and Pandey, Abhinav and Kadian, Abhishek and Al-Dahle, Ahmad and Letman, Aiesha and Mathur, Akhil and Schelten, Alan and Vaughan, Alex and others},
  journal={arXiv preprint arXiv:2407.21783},
  year={2024}
}

@article{grill2015black,
  title={Black-box optimization of noisy functions with unknown smoothness},
  author={Grill, Jean-Bastien and Valko, Michal and Munos, R{\'e}mi},
  journal={Advances in Neural Information Processing Systems},
  volume={28},
  year={2015}
}

@misc{haim2006second,
  title={The second {PASCAL} recognising textual entailment challenge},
  author={Haim, R Bar and Dagan, Ido and Dolan, Bill and Ferro, Lisa and Giampiccolo, Danilo and Magnini, Bernardo and Szpektor, Idan},
  year={2006}
}

@article{jamieson2012query,
  title={Query complexity of derivative-free optimization},
  author={Jamieson, Kevin G and Nowak, Robert and Recht, Ben},
  journal={Advances in Neural Information Processing Systems},
  volume={25},
  year={2012}
}

@inproceedings{ji2019improved,
  title={Improved zeroth-order variance reduced algorithms and analysis for nonconvex optimization},
  author={Ji, Kaiyi and Wang, Zhe and Zhou, Yi and Liang, Yingbin},
  booktitle={International Conference on Machine Learning},
  pages={3100--3109},
  year={2019},
  organization={PMLR}
}

@article{kingma2014adam,
  title={Adam: A method for stochastic optimization},
  author={Kingma, Diederik P and Ba, Jimmy},
  journal={arXiv preprint arXiv:1412.6980},
  year={2014}
}

@inproceedings{li2024addax,
  title={Addax: Utilizing Zeroth-Order Gradients to Improve Memory Efficiency and Performance of {SGD} for Fine-Tuning Language Models},
  author={Zeman Li and Xinwei Zhang and Peilin Zhong and Yuan Deng and Meisam Razaviyayn and Vahab Mirrokni},
  booktitle={International Conference on Learning Representations},
  year={2025}
}

@article{lin2022gradient,
  title={Gradient-free methods for deterministic and stochastic nonsmooth nonconvex optimization},
  author={Lin, Tianyi and Zheng, Zeyu and Jordan, Michael},
  journal={Advances in Neural Information Processing Systems},
  volume={35},
  pages={26160--26175},
  year={2022}
}

@inproceedings{liu2018signsgd,
  title={{S}ign{SGD} via Zeroth-Order Oracle},
  author={Sijia Liu and Pin-Yu Chen and Xiangyi Chen and Mingyi Hong},
  booktitle={International Conference on Learning Representations},
  year={2019},
}

@article{liu2018zeroth,
  title={Zeroth-order stochastic variance reduction for nonconvex optimization},
  author={Liu, Sijia and Kailkhura, Bhavya and Chen, Pin-Yu and Ting, Paishun and Chang, Shiyu and Amini, Lisa},
  journal={Advances in Neural Information Processing Systems},
  volume={31},
  year={2018}
}

@article{liu2019roberta,
  title={{RoBERTa}: A robustly optimized {BERT} pretraining approach},
  author={Liu, Yinhan and Ott, Myle and Goyal, Naman and Du, Jingfei and Joshi, Mandar and Chen, Danqi and Levy, Omer and Lewis, Mike and Zettlemoyer, Luke and Stoyanov, Veselin},
  journal={arXiv preprint arXiv:1907.11692},
  year={2019}
}

@article{liu2020primer,
  title={A primer on zeroth-order optimization in signal processing and machine learning: Principals, recent advances, and applications},
  author={Liu, Sijia and Chen, Pin-Yu and Kailkhura, Bhavya and Zhang, Gaoyuan and Hero III, Alfred O and Varshney, Pramod K},
  journal={IEEE Signal Processing Magazine},
  volume={37},
  number={5},
  pages={43--54},
  year={2020},
  publisher={IEEE}
}

@article{liu2024sparse,
  title={Sparse {M}e{ZO}: Less parameters for better performance in zeroth-order {LLM} fine-tuning},
  author={Liu, Yong and Zhu, Zirui and Gong, Chaoyu and Cheng, Minhao and Hsieh, Cho-Jui and You, Yang},
  journal={arXiv preprint arXiv:2402.15751},
  year={2024}
}

@inproceedings{ma2025revisiting,
  title={Revisiting Zeroth-Order Optimization: Minimum-Variance Two-Point Estimators and Directionally Aligned Perturbations},
  author={Ma, Shaocong and Huang, Heng},
  booktitle={International Conference on Learning Representations},
  year={2025}
}

@article{malladi2023fine,
  title={Fine-tuning language models with just forward passes},
  author={Malladi, Sadhika and Gao, Tianyu and Nichani, Eshaan and Damian, Alex and Lee, Jason D and Chen, Danqi and Arora, Sanjeev},
  journal={Advances in Neural Information Processing Systems},
  volume={36},
  pages={53038--53075},
  year={2023}
}

@article{mania2018simple,
  title={Simple random search of static linear policies is competitive for reinforcement learning},
  author={Mania, Horia and Guy, Aurelia and Recht, Benjamin},
  journal={Advances in Neural Information Processing Systems},
  volume={31},
  year={2018}
}

@book{nesterov2003introductory,
  title={Introductory lectures on convex optimization: A basic course},
  author={Nesterov, Yurii},
  volume={87},
  year={2003},
  publisher={Springer Science \& Business Media}
}

@article{nesterov2017random,
  title={Random gradient-free minimization of convex functions},
  author={Nesterov, Yurii and Spokoiny, Vladimir},
  journal={Foundations of Computational Mathematics},
  volume={17},
  pages={527--566},
  year={2017},
  publisher={Springer}
}

@article{park2025unraveling,
  title={Unraveling Zeroth-Order Optimization through the Lens of Low-Dimensional Structured Perturbations},
  author={Park, Sihwan and Yun, Jihun and Kim, SungYub and Kundu, Souvik and Yang, Eunho},
  journal={arXiv preprint arXiv:2501.19099},
  year={2025}
}

@inproceedings{rajpurkar2016squad,
  title={{SQuAD}: 100,000+ Questions for Machine Comprehension of Text},
  author={Rajpurkar, Pranav and Zhang, Jian and Lopyrev, Konstantin and Liang, Percy},
  booktitle={Proceedings of the Conference on Empirical Methods in Natural Language Processing},
  pages={2383--2392},
  year={2016}
}

@article{salimans2017evolution,
  title={Evolution strategies as a scalable alternative to reinforcement learning},
  author={Salimans, Tim and Ho, Jonathan and Chen, Xi and Sidor, Szymon and Sutskever, Ilya},
  journal={arXiv preprint arXiv:1703.03864},
  year={2017}
}

@article{shamir2017optimal,
  title={An optimal algorithm for bandit and zero-order convex optimization with two-point feedback},
  author={Shamir, Ohad},
  journal={Journal of Machine Learning Research},
  volume={18},
  number={1},
  pages={1703--1713},
  year={2017},
  publisher={JMLR}
}

@inproceedings{socher2013recursive,
  title={Recursive deep models for semantic compositionality over a sentiment treebank},
  author={Socher, Richard and Perelygin, Alex and Wu, Jean and Chuang, Jason and Manning, Christopher D and Ng, Andrew Y and Potts, Christopher},
  booktitle={Proceedings of the Conference on Empirical Methods in Natural Language Processing},
  pages={1631--1642},
  year={2013}
}

@article{spall1992multivariate,
  title={Multivariate stochastic approximation using a simultaneous perturbation gradient approximation},
  author={Spall, James C},
  journal={IEEE transactions on automatic control},
  volume={37},
  number={3},
  pages={332--341},
  year={1992},
  publisher={IEEE}
}

@article{team2024gemma,
  title={Gemma: Open models based on {G}emini research and technology},
  author={Team, Gemma and Mesnard, Thomas and Hardin, Cassidy and Dadashi, Robert and Bhupatiraju, Surya and Pathak, Shreya and Sifre, Laurent and Rivi{\`e}re, Morgane and Kale, Mihir Sanjay and Love, Juliette and others},
  journal={arXiv preprint arXiv:2403.08295},
  year={2024}
}

@article{team2024gemma2,
  title={Gemma 2: Improving open language models at a practical size},
  author={Team, Gemma and Riviere, Morgane and Pathak, Shreya and Sessa, Pier Giuseppe and Hardin, Cassidy and Bhupatiraju, Surya and Hussenot, L{\'e}onard and Mesnard, Thomas and Shahriari, Bobak and Ram{\'e}, Alexandre and others},
  journal={arXiv preprint arXiv:2408.00118},
  year={2024}
}

@article{team2025gemma,
  title={Gemma 3 technical report},
  author={Team, Gemma and Kamath, Aishwarya and Ferret, Johan and Pathak, Shreya and Vieillard, Nino and Merhej, Ramona and Perrin, Sarah and Matejovicova, Tatiana and Ram{\'e}, Alexandre and Rivi{\`e}re, Morgane and others},
  journal={arXiv preprint arXiv:2503.19786},
  year={2025}
}

@misc{tieleman2012lecture,
  title={Lecture 6.5 -- {RMSProp}: Divide the Gradient by a Running Average of Its Recent Magnitude},
  author={Tieleman, Tijmen and Hinton, Geoffrey},
  year={2012},
  howpublished={COURSERA: Neural Networks for Machine Learning}
}

@article{touvron2023llama,
  title={{LLaMA}: Open and efficient foundation language models},
  author={Touvron, Hugo and Lavril, Thibaut and Izacard, Gautier and Martinet, Xavier and Lachaux, Marie-Anne and Lacroix, Timoth{\'e}e and Rozi{\`e}re, Baptiste and Goyal, Naman and Hambro, Eric and Azhar, Faisal and others},
  journal={arXiv preprint arXiv:2302.13971},
  year={2023}
}

@article{touvron2023llama2,
  title={{LLAMA} 2: Open foundation and fine-tuned chat models},
  author={Touvron, Hugo and Martin, Louis and Stone, Kevin and Albert, Peter and Almahairi, Amjad and Babaei, Yasmine and Bashlykov, Nikolay and Batra, Soumya and Bhargava, Prajjwal and Bhosale, Shruti and others},
  journal={arXiv preprint arXiv:2307.09288},
  year={2023}
}

@inproceedings{voorhees2000building,
  title={Building a question answering test collection},
  author={Voorhees, Ellen M and Tice, Dawn M},
  booktitle={Proceedings of the Annual International ACM SIGIR Conference on Research and Development in Information Retrieval},
  pages={200--207},
  year={2000}
}

@inproceedings{wang2018glue,
  title={{GLUE}: A Multi-Task Benchmark and Analysis Platform for Natural Language Understanding},
  author={Wang, Alex and Singh, Amanpreet and Michael, Julian and Hill, Felix and Levy, Omer and Bowman, Samuel R},
  booktitle={International Conference on Learning Representations},
  year={2018}
}

@inproceedings{wang2018stochastic,
  title={Stochastic zeroth-order optimization in high dimensions},
  author={Wang, Yining and Du, Simon and Balakrishnan, Sivaraman and Singh, Aarti},
  booktitle={International Conference on Artificial Intelligence and Statistics},
  pages={1356--1365},
  year={2018},
  organization={PMLR}
}

@article{wang2022zeroth,
  title={Zeroth-order algorithms for nonconvex--strongly-concave minimax problems with improved complexities},
  author={Wang, Zhongruo and Balasubramanian, Krishnakumar and Ma, Shiqian and Razaviyayn, Meisam},
  journal={Journal of Global Optimization},
  pages={1--32},
  year={2022},
  publisher={Springer}
}

@article{wang2024simultaneous,
  title={Simultaneous Computation and Memory Efficient Zeroth-Order Optimizer for Fine-Tuning Large Language Models},
  author={Wang, Fei and Shen, Li and Ding, Liang and Xue, Chao and Liu, Ye and Ding, Changxing},
  journal={arXiv preprint arXiv:2410.09823},
  year={2024}
}

@article{wibisono2012finite,
  title={Finite sample convergence rates of zero-order stochastic optimization methods},
  author={Wibisono, Andre and Wainwright, Martin J and Jordan, Michael and Duchi, John C},
  journal={Advances in Neural Information Processing Systems},
  volume={25},
  year={2012}
}

@inproceedings{williams2018broad,
  title={A Broad-Coverage Challenge Corpus for Sentence Understanding through Inference},
  author={Williams, Adina and Nangia, Nikita and Bowman, Samuel},
  booktitle={Proceedings of the Conference of the North American Chapter of the Association for Computational Linguistics},
  pages={1112--1122},
  year={2018}
}

@inproceedings {xu2023federated,
  author={Mengwei Xu and Dongqi Cai and Yaozong Wu and Xiang Li and Shangguang Wang},
  title={{FwdLLM}: Efficient Federated Finetuning of Large Language Models with Perturbed Inferences},
  booktitle={USENIX Annual Technical Conference},
  year={2024},
  pages = {579--596}
}

@inproceedings{yue2023zeroth,
  title={Zeroth-order Optimization with Weak Dimension Dependency},
  author={Yue, Pengyun and Yang, Long and Fang, Cong and Lin, Zhouchen},
  booktitle={Conference on Learning Theory},
  pages={4429--4472},
  year={2023},
  organization={PMLR}
}

@article{zelikman2023just,
  title={Just One Byte (per gradient): A Note on Low-Bandwidth Decentralized Language Model Finetuning Using Shared Randomness},
  author={Zelikman, Eric and Huang, Qian and Liang, Percy and Haber, Nick and Goodman, Noah D},
  journal={arXiv preprint arXiv:2306.10015},
  year={2023}
}

@article{zhang2022opt,
  title={{OPT}: Open pre-trained {T}ransformer language models},
  author={Zhang, Susan and Roller, Stephen and Goyal, Naman and Artetxe, Mikel and Chen, Moya and Chen, Shuohui and Dewan, Christopher and Diab, Mona and Li, Xian and Lin, Xi Victoria and others},
  journal={arXiv preprint arXiv:2205.01068},
  year={2022}
}

@inproceedings{zhang2024dpzero,
  title={{DPZero}: Private Fine-Tuning of Language Models without Backpropagation},
  author={Zhang, Liang and Li, Bingcong and Thekumparampil, Kiran Koshy and Oh, Sewoong and He, Niao},
  booktitle={International Conference on Machine Learning},
  pages={59210--59246},
  year={2024},
  organization={PMLR}
}

@inproceedings{zhang2024revisiting,
  title={Revisiting Zeroth-Order Optimization for Memory-Efficient {LLM} Fine-Tuning: A Benchmark},
  author={Zhang, Yihua and Li, Pingzhi and Hong, Junyuan and Li, Jiaxiang and Zhang, Yimeng and Zheng, Wenqing and Chen, Pin-Yu and Lee, Jason D and Yin, Wotao and Hong, Mingyi and others},
  booktitle={International Conference on Machine Learning},
  pages={59173--59190},
  year={2024},
  organization={PMLR}
}

@article{zhang2025zeroth,
  title={Zeroth-Order Optimization Finds Flat Minima},
  author={Zhang, Liang and Li, Bingcong and Thekumparampil, Kiran Koshy and Oh, Sewoong and Muehlebach, Michael and He, Niao},
  journal={arXiv preprint arXiv:2506.05454},
  year={2025}
}

@inproceedings{zhao2024second,
  title={Second-Order Fine-Tuning without Pain for {LLM}s: A {H}essian Informed Zeroth-Order Optimizer},
  author={Yanjun Zhao and Sizhe Dang and Haishan Ye and Guang Dai and Yi Qian and Ivor Tsang},
  booktitle={International Conference on Learning Representations},
  year={2025}
}

@inproceedings{pilehvar2019wic,
  title={{WiC}: The Word-in-Context Dataset for Evaluating Context-Sensitive Meaning Representations},
  author={Pilehvar, Mohammad Taher and Camacho-Collados, Jose},
  booktitle={Proceedings of the Conference of the North American Chapter of the Association for Computational Linguistics},
  pages={1267--1273},
  year={2019}
}

@article{zhang2018record,
  title={{ReCoRD}: Bridging the gap between human and machine commonsense reading comprehension},
  author={Zhang, Sheng and Liu, Xiaodong and Liu, Jingjing and Gao, Jianfeng and Duh, Kevin and Van Durme, Benjamin},
  journal={arXiv preprint arXiv:1810.12885},
  year={2018}
}

@inproceedings{khashabi2018looking,
  title={Looking beyond the surface: A challenge set for reading comprehension over multiple sentences},
  author={Khashabi, Daniel and Chaturvedi, Snigdha and Roth, Michael and Upadhyay, Shyam and Roth, Dan},
  booktitle={Proceedings of the Conference of the North American Chapter of the Association for Computational Linguistics},
  pages={252--262},
  year={2018}
}

\clearpage
\clearpage

\appendix
\crefalias{section}{appendix}
\crefalias{subsection}{appendix}
\crefalias{subsubsection}{appendix}
\renewcommand{\sectionautorefname}{Appendix}
\renewcommand{\subsectionautorefname}{Appendix}
\renewcommand{\subsubsectionautorefname}{Appendix}
\onecolumn
\aistatstitle{ConMeZO: Adaptive Descent-Direction Sampling for Gradient-Free Finetuning of Large Language Models: \\
Supplementary Material}

\section{PROOFS}

\subsection{Proof of \cref{prop:orth}}
\label{app:proof_orth}

Let $u_t \sim \mathcal U(\mathbb S ^{d-1})$ and $\hat m_t \in \mathbb R ^d$ with $\| \hat m_t \| = 1$ be a promising search direction. Instead of ensuring that the sampled random direction is orthogonal to $\hat m_t$, we show that it suffices to sample any random direction $u_t \sim \mathcal U(\mathbb S ^{d-1})$. We show that the relative magnitude of the projection $\langle \hat m_t, u_t \rangle$ becomes negligible as $d \to \infty$.

\begin{Proof}
    \label{app:p1}

Notice that $u_t$ can be sampled as $u_t = \frac{X}{\| X \|}$, where $X \sim \mathcal N(0, I_d)$.
We have that

$$
{\l < \hat m_t , u_t \r >} = \frac{\l< \hat m_t, X\r>}{\|X\|}.
$$

$\l < \hat m_t , X \r >$ is a $\mathcal N(0,1)$ random variable since $\| \hat m_t \| = 1$, and $\|X\|^2$ is a $\chi^2$-distributed random variable with $d$ degrees of freedom. Therefore, for large $d$, the ratio $\l < \hat m_t , X \r >/\|X\|$ is on the order of $\mathcal N(0,1)/ \sqrt d$, which converges to $0$ in probability as $d \to \infty$.

\end{Proof}

\subsection{Proof of \cref{prop:angle}}
\label{app:proof_angle}

Consider a cone $ \mathcal{C} $ in $ \mathbb{R}^d $ with apex at the origin, central axis aligned with a unit vector $ \hat m_t $, and half-angle $ \theta \in [0, \pi /2]$. We are interested in the distribution of the angle $ \gamma $ between a random vector $ z_t $, sampled uniformly from the intersection of $  \mathcal{C} $ with $\sqrt{d} \,\mathbb S^{d-1}$, and the axis $ \hat m_t $. The following proof demonstrates that as the dimension $ d \to \infty $, the angle $ \gamma $ becomes concentrated at $ \theta $.

\begin{Proof}

We consider the case where $d \to \infty$:

\begin{align*}
    p(\gamma \leq \theta ') & = \frac{\int_0^{\theta'} \text{ Surface area of hypercircle with radius }r( \alpha) \ d\alpha}{\int_0^{\theta}\text{Surface area of hypercircle with radius }r(\beta) \ d\beta} \\
    & = \frac{\int_0^{\theta'} C_d\cdot r(\alpha)^{d-1} \ d \alpha}{\int_0^{\theta} C_d \cdot r (\beta)^{d-1} \ d \beta}
\end{align*}

where $C_d$ is a constant dependent on $d$ and independent of radius.
When inspecting Figure \ref{fig:cone3}, it is simple to see that $r(\gamma) = \sqrt{d} \sin (\gamma)$. Now assume that $\theta ' < \theta$. Further calculation yields
\begin{align*}
    p(\gamma \leq \theta') & = \frac{\int_0^{\theta'} (\sqrt{d} \sin(\alpha))^{d-1}\ d \alpha}{\int_0^\theta  (\sqrt{d} \sin(\beta))^{d-1}\ d \beta} \\
    & \leq \frac{\theta' (\sin (\theta'))^{d-1}}{\int_0^\theta  (\sin(\beta))^{d-1}\ d \beta}
\end{align*}
because $(\sin \alpha)^{d-1} \leq (\sin \theta')^{d-1}$ for $ 0 \leq \alpha \leq \theta'$, and the interval length is $\theta'$. Let $s=\frac{\theta + \theta '}{2} \in (\theta',\theta)$. Now because $\sin (\beta)$ is increasing on $[0, \theta]$, on the sub-interval $[s, \theta]$ we have $(\sin (\beta))^{d-1} \geq (\sin (s))^{d-1}$.
Hence
$$
\int_{0}^{\theta}(\sin (\beta))^{d-1}\,d\beta \geq
\int_{s}^{\theta} (\sin (\beta))^{d-1} \,d\beta
\geq
(\theta-s) \, (\sin(s))^{d-1}.
$$
We have that
$$
\sin (s) > \sin (\theta'),
$$
since $\theta > s > \theta'$.
Putting these together, we get

\begin{align*}
    p(\gamma \leq \theta') & = \frac{\int_0^{\theta'} (\sin(\alpha))^{d-1}\ d \alpha}{\int_0^\theta  (\sin(\beta))^{d-1}\ d \beta} \\
    & \leq \frac {\theta'}{\theta - s} \left (\frac{\sin (\theta ')}{ \sin (s)} \right )^{d-1}.
\end{align*}

Since $\sin (s) > \sin ( \theta')$, we have that

$$
p(\gamma \leq \theta') \to 0  \text{ for } d \to \infty .
$$

So instead of sampling $\gamma$, in practice we can set $\gamma= \theta$.

\label{app:p3}

\end{Proof}

\subsection{Proof of \cref{lem:first_moment}}
\label{app:proof_first_moment}

\begin{Proof}
We have that $z_t = \sqrt{d} \cos (\theta) \cdot \hat{m}_t + \sin ( \theta ) \cdot u_t$, where $u_t \sim \mathcal{U}( \sqrt{d} \,\mathbb S^{d-1})$.
\begin{align*}
\bE_{u_t} \l [ (z_t ^\top a_t) z_t \r] &= 
\bE_{u_t} \l [ \left( \left(\cos ( \theta ) \sqrt d \cdot \hat m_t + \sin ( \theta ) \cdot u_t \right)^\top a_t \right)(\cos ( \theta ) \sqrt d \cdot \hat m_t + \sin ( \theta ) \cdot u_t)\r] \\
&= d\cos^2 ( \theta ) (\hat m_t^\top a_t) \cdot \hat m_t + \sin^2 ( \theta ) \cdot \mathbb E \l [
(u_t^\top a_t)u_t
\r] \\
&= d \cos^2 ( \theta ) (\hat m_t^\top a_t) \cdot \hat m_t + \sin^2 ( \theta ) \cdot a_t.
\end{align*}

Let $z_t=\alpha \hat m_t + \beta u_t$, where $u_t \sim \mathcal U(\sqrt d \ \mathbb S^{d-1})$ , $\|\hat m_t\|=1$, $\alpha = \cos (\theta) \cdot \sqrt d$ and $\beta = \sin (\theta)$.
We now derive the second moment of $(z_t^\top a_t)z_t$.

    \begin{align*}
        & \bE_{u_t} \l [ \l \| (z_t^\top a_t)z_t \r \| ^2 \r ] \\
        = \quad &\bE_{u_t} \l[\l \| \l(\alpha (\hat m_t^\top a_t) + \beta (u_t^\top a_t)\r)(\alpha \hat m_t + \beta u_t) \r\|^2\r] \\
        = \quad & \bE_{u_t} \l[ \l( \alpha^2 (\hat m_t^\top a_t)^2 + \beta^2 (u_t^\top a_t)^2 + 2 \alpha \beta (\hat m_t^\top a_t)(u_t^\top a_t)\r)  \l ( \alpha^2 \| \hat m_t \|^2 + \beta^2 \|u_t \|^2 + 2 \alpha \beta (\hat m_t^\top u_t)\r)\r] \\
        = \quad & \alpha ^2 \left ( \alpha ^2 (\hat m_t^\top a_t)^2 +  \beta^2 \|a_t\|^2 + 0 \right ) \\
        + & \beta^2 \left ( \alpha^2 (\hat m_t^\top a_t)^2 d + \beta^2 d \|a_t\|^2 + 0\right) \\
        + & 2 \alpha \beta \l ( 0+0+2 \alpha \beta (\hat m_t^\top a_t)^2\r) \\
        = \quad & d^2 \cos^2 (\theta) (\hat m_t^\top a_t)^2 + 4d \sin^2 (\theta) \cos^2 (\theta) (\hat m_t^\top a_t)^2 + d \sin^2 (\theta) \|a_t\|^2 \\
        = \quad & d \cos^2 (\theta) \l( d+4 \sin^2 (\theta) \r) (\hat m_t^\top a_t)^2 + d \sin^2 (\theta) \|a_t \|^2 \\
        \leq \quad & d (d+4) \cos^2 (\theta) (\hat m_t^\top a_t)^2 + d \sin^2 (\theta) \| a_t \|^2.
    \end{align*}

    Using $(\hat m_t^\top a_t)^2 = (\cos (\rho) \| \hat m_t \| \| a_t \|)^2 = \cos^2 (\rho) \cdot \| a_t \| ^2$, we have:
    \begin{align*}
    \bE_{u_t} \l [ \l \| (z_t^\top a_t)z_t \r \| ^2 \r ] & \leq d \left((d+4) \cos^2 (\theta) \cos^2 (\rho) + \sin^2 (\theta) \right) \|a_t \|^2.
    \end{align*}
\end{Proof}

\subsection{Proof of \Cref{thm:improvement_per_it}}
\label{app:proof_improvement_per_it}

We assume that the function $f$ is $\ell$-smooth, meaning:
$$
f(x_{t+1}) \leq f(x_t) + \nabla f(x_t)^\top (x_{t+1} - x_t) + \frac{\ell}{2} \|x_{t+1} - x_t\|^2.
$$
Additionally, we use the update rule $x_{t+1} = x_t - \eta g_\lambda(x_t,z_t)$, where $g_\lambda(x_t,z_t)$ is the gradient estimate at $x_t$ and $z_t = \cos (\theta) \sqrt{d} \cdot \hat{m}_t + \sin ( \theta ) \cdot u_t$, where $u_t \sim \mathcal{U}(\sqrt d\, \mathbb S^{d-1})$. Let $a_t = \nabla f(x_t)$. The proof derives the expected improvement in $f(x)$ per iteration under these assumptions.

\begin{Proof}
Substituting the update rule $x_{t+1} = x_t - \eta g_\lambda(x_t,z_t)$ into the smoothness assumption, we have that
$$
f(x_{t+1}) \leq f(x_t) - \eta \nabla f(x_t)^\top g_\lambda(x_t,z_t) + \frac{\eta^2 \ell}{2} \|g_\lambda(x_t,z_t)\|^2.
$$
Taking expectations with respect to $u_t$, the random search direction, we obtain
$$
\bE_{u_t}[f(x_{t+1})] \leq f(x_t) - \eta a_t^\top \bE_{u_t}[g_\lambda(x_t,z_t)] + \frac{\eta^2 \ell}{2} \bE_{u_t}[\|g_\lambda(x_t,z_t)\|^2].
$$
Using the moments of $g_\lambda(x_t,z_t)$ with $\lambda\to0$:
$$
\bE_{u_t}[(z_t^\top a_t)z_t] = d \cos^2 (\theta) (\hat{m}_t^\top a_t) \cdot \hat{m}_t + \sin^2 (\theta) \cdot  a_t,
$$
and
$$
\bE_{u_t}\l[\|(z_t^\top a_t)z_t\|^2\r] \leq d \left((d+4) \cos^2 (\theta) \cos^2 (\rho) + \sin^2 (\theta)\right) \|a_t\|^2,
$$
we substitute these into the inequality and get that
\begin{align*}
\bE_{u_t}[f(x_{t+1})] &\leq f(x_t) - \eta \left(d \cos^2 (\theta) \cos^2 (\rho) + \sin^2 (\theta) \right) \|a_t\|^2 \\
&\quad + \frac{\eta^2 \ell}{2} d \left((d+4) \cos^2 (\theta) \cos^2 (\rho) + \sin^2 (\theta) \right) \|a_t\|^2.
\end{align*}
This proves the first part of the theorem.

Next, rearranging the inequality to isolate \(\|a_t\|^2\), we have that
\begin{align*}
\bE_{u_t}[f(x_t) - f(x_{t+1})] & \geq \Big(\eta \left(d \cos^2 (\theta) \cos^2 (\rho) + \sin^2 (\theta) \right) \\
&\quad - \frac{\eta^2 \ell}{2} d \left((d+4) \cos^2 (\theta) \cos^2 (\rho) + \sin^2 (\theta)\right)\Big) \|a_t\|^2.
\end{align*}
Thus, it holds that
$$
\|a_t\|^2 \leq \frac{\bE_{u_t}[f(x_t) - f(x_{t+1})]}{\eta \left(d \cos^2 (\theta) \cos^2 (\rho) + \sin^2 (\theta) \right) - \frac{\eta^2 \ell}{2} d \left((d+4) \cos^2 (\theta) \cos^2 (\rho) + \sin^2 (\theta) \right)}.
$$

Observe that the denominator is a concave quadratic function of \(\eta\) and achieves its maximum at:
$$
\eta^* = \frac{d \cos^2 (\theta) \cos^2 (\rho) + \sin^2 (\theta)}{\ell d \left((d+4) \cos^2 (\theta) \cos^2 (\rho) + \sin^2 (\theta)\right)}.
$$

Substituting \(\eta = \eta^*\) and rearranging for $\bE_{u_t}[f(x_{t+1})]-f(x_t)$ yields:

\begin{align*}
\bE_{u_t}[f(x_{t+1})]-f(x_t) & \leq - \left ( \eta^* \left(d \cos^2 (\theta) \cos^2 (\rho) + \sin^2 (\theta) \right)  - \frac{{\eta^*}^2 \ell}{2} d \left((d+4) \cos^2 (\theta) \cos^2 (\rho) + \sin^2 (\theta) \right) \right )\| a_t \|^2 \\
& = - \frac{\left(d \cos^2 (\theta) \cos^2 (\rho) + \sin^2 (\theta)\right)^2}{2 \ell d \left((d+4) \cos^2 (\theta) \cos^2 (\rho) + \sin^2 (\theta)\right)} \|a_t\|^2 \\
& \leq - \frac{\left(d \cos^2 (\theta) \cos^2 (\rho) + \sin^2 (\theta)\right)^2}{2 \ell d \left((d+4) \cos^2 (\theta) \cos^2 (\rho) + \frac{d+4}{d} \sin^2 (\theta)\right)} \|a_t\|^2 \\
& = - \frac{d \cos^2 (\theta) \cos^2 (\rho) + \sin^2 (\theta)}{2 \ell d (1+4/d)}  \|a_t\|^2 \\
& \approx- \frac{d \cos^2 (\theta) \cos^2 (\rho) + \sin^2 (\theta)}{2 \ell d }  \|a_t\|^2.
\end{align*}
This completes the proof of the theorem.

\end{Proof}
\section{IMPLEMENTATION AND PRACTICAL SPEEDUPS}\label{sec:speedups}

As an additional contribution, ConMeZO introduces an efficient implementation framework that reduces wall-clock training time. This speedup arises from two complementary effects:  
(1) the optimizer converges in fewer iterations due to more informative, low-variance search directions, and  
(2) each iteration itself is faster thanks to a fully vectorized implementation.

Most MeZO variants retain similar algorithmic structure and perform perturbations and updates inside Python loops, generating random tensors for each parameter separately. In contrast, ConMeZO maintains a single flattened parameter buffer and applies perturbations or updates through fused in-place vector operations. This avoids repeated kernel launches, reduces RNG overhead, and minimizes Python-side iteration costs.

Beyond ConMeZO itself, this implementation framework is broadly applicable to other momentum-based zeroth-order optimizers, offering a general recipe for improving their computational efficiency without modifying the underlying algorithmic logic.

\lstset{
  basicstyle=\ttfamily\small,
  breaklines=true,
  frame=single,
  language=Python,
  showstringspaces=false,
  columns=fullflexible,
  xleftmargin=4pt,
  xrightmargin=4pt
}

\noindent\textbf{MeZO (loop-based perturbation):}
\begin{lstlisting}
def efficient_perturb_parameters(self, model, random_seed, scaling_factor=1):
    torch.manual_seed(random_seed)
    for name, param in self.named_parameters_to_optim:
        u = torch.normal(mean=0, std=1, size=param.shape, device=param.device)
        param.data += scaling_factor * u * self.args.zero_order_eps
    return model
\end{lstlisting}

\noindent\textbf{ConMeZO (vectorized perturbation):}
\begin{lstlisting}
def efficient_perturb_parameters(self, model, random_seed, scaling_factor=1):
    fac = np.sqrt(self.d) / self.momentum_norm
    alpha = fac * np.cos(self.args.cone_theta)
    self.params_flat.add_(self.momentum_flat,
        alpha=scaling_factor * alpha * self.args.zero_order_eps)
    return model
\end{lstlisting}

For reference, ConMeZO samples the random direction once and stores the full perturbation in the momentum buffer. This design allows all three perturbations and the model update to be performed in a single vectorized pass, whereas MeZO re-samples random directions four times.

While ConMeZO improves computational efficiency, it introduces a modest and constant memory overhead due to maintaining the momentum vector. As shown in Table~\ref{tab:mem_combined}, this results in a consistent increase in peak memory usage across models and tasks.

\begin{table*}
\centering
\caption{ConMeZO increases peak memory usage (MiB) after 100 steps by a constant amount for each model across all the tasks}
\label{tab:mem_combined}
\setlength{\tabcolsep}{4pt}
\resizebox{\columnwidth}{!}{
\begin{tabular}{lcccccccccccc}
\toprule
 & \multicolumn{7}{c}{RoBERTa-Large} & \multicolumn{5}{c}{OPT-1.3B} \\
\cmidrule(lr){2-8} \cmidrule(lr){9-13}
Method & SST2 & SST5 & SNLI & MNLI & RTE & TREC & Avg. & SST2 & BoolQ & DROP & SQuAD & Avg. \\
\midrule
MeZO     & \textbf{2178} & \textbf{2178} & \textbf{2177} & \textbf{2549} & \textbf{2542} & \textbf{2178} & \textbf{2300} & \textbf{3510}  & \textbf{8983}  & \textbf{24915} & \textbf{8621} & \textbf{11507} \\
ConMeZO  & 3343 & 3343 & 3343 & 4105 & 4099 & 3343 & 3596 & 6020  & 11493 & 27425 & 11132 & 14017 \\
\midrule
$\Delta$ Increase & 1165 & 1165 & 1165 & 1556 & 1557 & 1165 & 1295 & 2510 & 2510 & 2511 & 2510 & 2510 \\
\% Increase & 53.5\% & 53.5\% & 53.5\% & 61.0\% & 61.2\% & 53.5\% & 56.0\% & 71.5\% & 27.9\% & 10.1\% & 29.1\% & 34.6\% \\
\bottomrule
\end{tabular}
}
\end{table*}

\section{EXPERIMENTAL RESULTS AND DETAILS}
\label{app:expts_appendix}

\subsection{Synthetic Experiments}
\label{apdx.sec.synthetic}
This section provides details and additional results for the experiments in \Cref{sec:synthetic_experiments}.
Our objective is the quadratic problem $f(x) = \sum_{i=1}^d \sigma_i x_i^2$, where $(\sigma_i)$ is a geometric series with initial value $1.0/d$ and final value $1.0$. Hence the problem is strongly convex with condition number $d$. The initial iterate $x_0$ is randomly sampled from the vectors with norm $10.0$. We set $\lambda = 0.01$ and tune the rest of the hyperparameters on the grid with choices $\eta = \{10^0, 10^{-1}, 10^{-2}, 10^{-3}, 10^{-4}\}$, $\beta = \{0.8, 0.9, 0.95, 0.99\}$, $\theta = \{1.2, 1.3, 1.4, 1.5\}$. Tuning is done over $100,000$ iterations and the final plot is shown for $100,000$ iterations. Note that we do not apply any momentum parameter warm-up for synthetic experiments.

\subsection{RoBERTa}\label{apdx.sec.roberta}
This section provides details and additional results for the experiments in \Cref{sec:roberta_experiments}.

The finetuning process involved six standard NLP tasks from the GLUE benchmark \citep{wang2018glue}: SST-2 (binary sentiment classification), SST-5 (fine-grained sentiment classification) \citep{socher2013recursive}, SNLI (natural language inference) \citep{bowman2015large}, MNLI (multi-genre natural language inference) \citep{williams2018broad}, RTE (recognizing textual entailment) \citep{dagan2005pascal, haim2006second, giampiccolo2007third, bentivogli2009fifth, wang2018glue}, and TREC (question classification) \citep{voorhees2000building}.

The experiments were conducted by finetuning RoBERTa-large in a few-shot setting, using 512 samples per class, to evaluate the optimizer's performance.
All experiments are executed on a single NVIDIA H100 GPU with $\sim$95 GiB of memory. 
Our implementation samples random directions from a standard normal distribution instead of from the sphere. This common practice simplifies implementation without hurting performance \cite{zhang2024dpzero}.
We fixed hyperparameters for our optimizer to $\theta = 1.35$ and $\beta = 0.99$ for reporting all results, while the learning rate $\eta = 10^{-6}$ and smoothing parameter $\lambda = 10^{-3}$ were fixed for both MeZO and ConMeZO to ensure a fair comparison. We tuned hyperparameters $\theta$ and $\beta$ on the grid with choices $\theta=\{1.35, 1.4\}$, $\beta = \{0.95, 0.99\}$.

\begin{figure}[!t]
  \centering
  \includegraphics[width=0.77\textwidth]{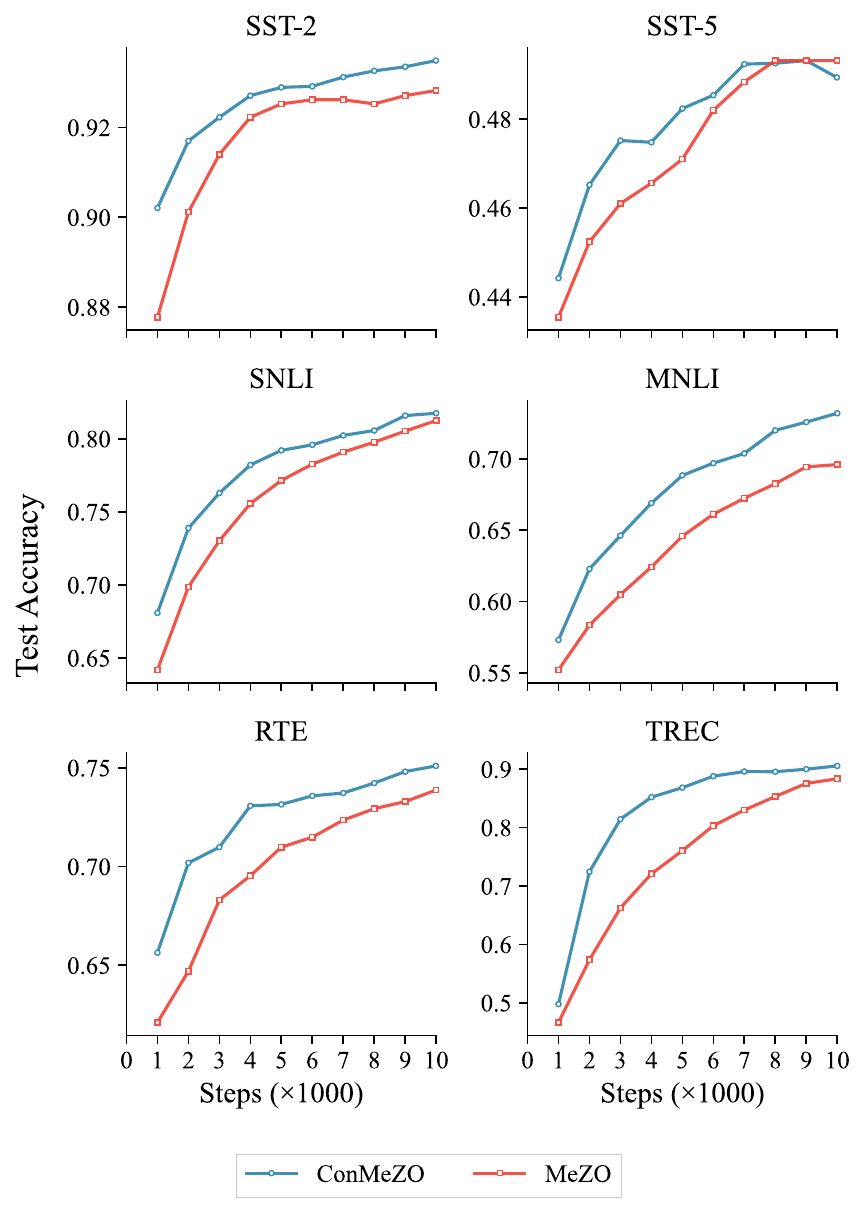}
  \caption{Test Accuracy of ConMeZO with settings mentioned in \ref{apdx.sec.roberta} compared to MeZO over 10,000 iterations.}
  \label{fig:test_acc_10k}
\end{figure}

Our implementation builds on the framework provided by DPZero paper \citep{zhang2024dpzero} and can be found at \url{https://github.com/LejsDeen/ConMeZO}.
The optimizer’s performance is analyzed under varying configurations of its hyperparameters: $\theta$, $\beta$, and the learning rate $\eta$. These parameters are systematically adjusted to evaluate their sensitivity and impact on model performance.
The smoothing parameter $\lambda$ is fixed to $10^{-3}$ for all experiments.
We use the seeds $13, 21, 42, 87, 100$ to calculate values provided in \Cref{tab:performance_10k,tab:performance_10k_std,tab:acc_comparison_dev}. In \Cref{fig:test_acc_10k} we provide test accuracy curves for a direct comparison between MeZO and ConMeZO.

\paragraph{Comparison to First-Order SGD.}
For completeness, we additionally report results for standard first-order SGD (FO-SGD) on two representative tasks, SST-2 and RTE in Table~\ref{tab:performance_10k_sgd}. We use the test accuracies reported in \citet{zhang2024revisiting} for FO-SGD. 

\begin{table}[]
\centering
\caption{RoBERTa-Large test performance. We include standard SGD as an additional FO baseline; remarkably, ConMeZO can outperform SGD on tasks like RTE.}
\begin{tabular}{lccccc}
\toprule
& \multicolumn{2}{c}{FO Methods} & \multicolumn{3}{c}{ZO Methods} \\
\cmidrule(lr){2-3} \cmidrule(lr){4-6}
Task   & AdamW & SGD & MeZO & Mom. & ConMeZO \\
\midrule
SST-2  & 93.1 & 91.6 & 92.8 & 92.2 & \textbf{93.5} \\
RTE    & 83.6 & 70.9 & 73.9 & 74.0 & \textbf{75.1} \\
\bottomrule
\end{tabular}
\label{tab:performance_10k_sgd}
\end{table}

\paragraph{Parameter Sensitivity \& Ablation Study.}
\label{chap:ablation}

Understanding the sensitivity of the optimizer to its hyperparameters, particularly momentum ($\beta$) and cone angle ($\theta$), provides critical insights into its performance across different phases of optimization. Section \ref{additional_results} explores their roles in convergence acceleration and alignment with the true gradient, highlighting key patterns observed in the experiments.

\begin{table}[]
  \setlength{\tabcolsep}{4.5pt}
  \centering
  \caption{Final test accuracy (\%) of RoBERTa Large after 10{,}000 iterations with MeZO and ConMeZO, averaged over 5 seeds. Entries are mean $\pm$ std across seeds.}
  \begin{tabular}{lccccccc}
    \toprule
    & SST-2 & SST-5 & SNLI & MNLI & RTE & TREC & Avg. \\
    \midrule
    MeZO     & 92.8 $\pm$ 0.6 & \textbf{49.3} $\pm$ 1.1 & 81.0 $\pm$ 0.4 & 69.7 $\pm$ 0.9 & 73.9 $\pm$ 0.9 & 88.4 $\pm$ 1.8 & 75.85 \\
    ConMeZO  & \textbf{93.5} $\pm$ 0.7 & 48.9 $\pm$ 0.9 & \textbf{81.9} $\pm$ 0.8 & \textbf{73.2} $\pm$ 1.2 & \textbf{75.1} $\pm$ 1.5 & \textbf{90.0} $\pm$ 0.7 & \textbf{77.11} \\
    \bottomrule
  \end{tabular}
  \label{tab:performance_10k_std}
\end{table}

\begin{table}[]

  \centering
  \caption{Accuracy (\%) with standard deviation of ConMeZO vs.\ MeZO on RoBERTa at different steps averaged over 5 seeds.}
  \label{tab:acc_comparison_dev}
  \begin{tabular}{c *{6}{c}}
    \toprule
    & \multicolumn{2}{c}{SST-2}
    & \multicolumn{2}{c}{SST-5}
    & \multicolumn{2}{c}{SNLI} \\
    \cmidrule(lr){2-3} \cmidrule(lr){4-5} \cmidrule(l){6-7}
    \textbf{Steps} & MeZO & ConMeZO & MeZO & ConMeZO & MeZO & ConMeZO \\
    \midrule
    1500 & 87.8 {\scriptsize$\pm$0.4} & \bfseries 90.2 {\scriptsize$\pm$1.1} & 
           43.5 {\scriptsize$\pm$1.0} & \bfseries 44.4 {\scriptsize$\pm$1.4} & 
           64.2 {\scriptsize$\pm$1.2} & \bfseries 68.1 {\scriptsize$\pm$1.3} \\
    3000 & 91.4 {\scriptsize$\pm$0.5} & \bfseries 92.2 {\scriptsize$\pm$0.7} & 
           46.1 {\scriptsize$\pm$1.0} & \bfseries 47.5 {\scriptsize$\pm$1.3} & 
           73.0 {\scriptsize$\pm$1.3} & \bfseries 76.3 {\scriptsize$\pm$1.4} \\
    6000 & 92.6 {\scriptsize$\pm$0.7} & \bfseries 92.9 {\scriptsize$\pm$0.7} & 
           48.2 {\scriptsize$\pm$0.9} & \bfseries 48.5 {\scriptsize$\pm$1.1} & 
           78.3 {\scriptsize$\pm$0.8} & \bfseries 79.6 {\scriptsize$\pm$0.3} \\
    \bottomrule
  \end{tabular}

  \vspace{8pt}

  \begin{tabular}{c *{6}{c}}
  
    \toprule
    & \multicolumn{2}{c}{MNLI}
    & \multicolumn{2}{c}{RTE}
    & \multicolumn{2}{c}{TREC} \\
    \cmidrule(lr){2-3} \cmidrule(lr){4-5} \cmidrule(l){6-7}
    \textbf{Steps} & MeZO & ConMeZO & MeZO & ConMeZO & MeZO & ConMeZO \\
    \midrule
    1500 & 55.2 {\scriptsize$\pm$0.2} & \bfseries 57.3 {\scriptsize$\pm$0.4} & 
           62.1 {\scriptsize$\pm$0.6} & \bfseries 65.6 {\scriptsize$\pm$1.5} & 
           46.6 {\scriptsize$\pm$2.1} & \bfseries 49.8 {\scriptsize$\pm$3.4} \\
    3000 & 60.5 {\scriptsize$\pm$0.6} & \bfseries 64.6 {\scriptsize$\pm$1.3} & 
           68.3 {\scriptsize$\pm$0.5} & \bfseries 71.0 {\scriptsize$\pm$0.7} & 
           66.2 {\scriptsize$\pm$2.5} & \bfseries 81.4 {\scriptsize$\pm$3.3} \\
    6000 & 66.1 {\scriptsize$\pm$0.5} & \bfseries 69.7 {\scriptsize$\pm$1.7} & 
           71.5 {\scriptsize$\pm$0.5} & \bfseries 73.6 {\scriptsize$\pm$0.6} & 
           80.3 {\scriptsize$\pm$2.0} & \bfseries 88.8 {\scriptsize$\pm$0.9} \\
    \bottomrule
  \end{tabular}
\end{table}

\begin{table}[t]
\centering
\caption{OPT 1.3B Test Accuracy or F1 score in \%. Mean $\pm$ std over 3 seeds.}
\setlength{\tabcolsep}{4pt}
\begin{tabular}{lcccc}
\toprule
& SQuAD & SST2 & WIC & BoolQ \\
\midrule
MeZO     & 72.76 $\pm$ 0.97 & 88.49 $\pm$ 0.29 & 56.53 $\pm$ 0.72 & 63.50 $\pm$ 0.78 \\
ConMeZO  & \textbf{75.34} $\pm$ 1.41 & \textbf{90.56} $\pm$ 0.83 & \textbf{58.15} $\pm$ 0.57 & \textbf{64.20} $\pm$ 1.90 \\
\midrule
& DROP & ReCoRD & RTE & MultiRC \\
\midrule
MeZO     & 25.90 $\pm$ 2.34 & 70.67 $\pm$ 0.75 & \textbf{56.92} $\pm$ 1.16 & \textbf{55.90} $\pm$ 1.23 \\
ConMeZO  & \textbf{26.53} $\pm$ 1.92 & 70.67 $\pm$ 0.45 & 55.48 $\pm$ 3.62 & 53.50 $\pm$ 1.41 \\
\midrule
\textbf{Average} & \multicolumn{2}{c}{MeZO: 61.33} & \multicolumn{2}{c}{ConMeZO: \textbf{61.80}} \\
\bottomrule
\end{tabular}
\label{tbl:opt_1.3b_std}
\end{table}

\begin{table}[t]
\centering
\caption{OPT 13B Test Accuracy or F1 score in \%. Mean $\pm$ std over 3 seeds.}
\begin{tabular}{lcccc}
\toprule
& SQuAD & SST2 & WIC & BoolQ \\
\midrule
MeZO     & 82.28 $\pm$ 1.02 & 91.25 $\pm$ 0.18 & \textbf{58.93} $\pm$ 0.57 & 67.73 $\pm$ 2.16 \\
ConMeZO  & \textbf{83.66} $\pm$ 0.71 & \textbf{92.39} $\pm$ 0.58 & 58.31 $\pm$ 1.66 & \textbf{69.33} $\pm$ 2.31 \\
\midrule
& DROP & ReCoRD & RTE & MultiRC \\
\midrule
MeZO     & OOM & \textbf{81.17} $\pm$ 1.04 & 63.66 $\pm$ 1.27 & 57.53 $\pm$ 2.27 \\
ConMeZO  & OOM & 80.87 $\pm$ 1.00 & \textbf{64.50} $\pm$ 1.63 & \textbf{58.30} $\pm$ 0.56 \\
\midrule
\textbf{Average} & \multicolumn{2}{c}{MeZO: 71.79} & \multicolumn{2}{c}{ConMeZO: \textbf{72.48}} \\
\bottomrule
\end{tabular}
\label{tbl:opt_13b_std}
\end{table}

\subsection{OPT}\label{apdx.sec.opt}
This section provides details and additional results for the experiments in \Cref{sec:opt_experiments}.

All OPT-1.3B and 13B experiments are conducted for 20K iterations using both ConMeZO and MeZO with a fixed learning rate of $\eta = 10^{-7}$. Due to the high cost of finetuning a 1.3 and 13B-parameter model and rerunning multiple random seeds, additional learning-rate searches are not performed. Consequently, every OPT result reported use three seeds and $\eta=10^{-7}$ is adopted for both optimizers. These findings align well with our RoBERTa results, demonstrating that the observed performance gains under fixed hyperparameter settings extend across different model scales. 
We use seeds 0, 29, and 83 for our reported results.

Our implementation builds on the DPZero framework \citep{zhang2024dpzero}.
 We fix the smoothing parameter to $\lambda=10^{-3}$ and, instead of sampling uniformly from a hypersphere, draw random directions from $\mathcal{N}(0,I_d)$, which is a valid simplification in high dimensions.

 Fine‐tuning is conducted on decoder‐only Transformers OPT‐1.3B and OPT‐13B \citep{zhang2022opt}, respectively.
We evaluate ConMeZO on eight benchmarks spanning diverse reasoning and understanding skills:
SST‐2 \citep{socher2013recursive} (binary sentiment classification),
BoolQ \citep{clark2019boolq} (boolean question answering),
SQuAD v1.1 \citep{rajpurkar2016squad} (span‐based QA),
DROP \citep{dua2019drop} (discrete reasoning QA),
WiC \citep{pilehvar2019wic} (word sense disambiguation),
ReCoRD \citep{zhang2018record} (reading comprehension with commonsense reasoning),
RTE \citep{dagan2005pascal,bentivogli2009fifth} (textual entailment), and
MultiRC \citep{khashabi2018looking} (multi‐sentence reasoning and multiple‐choice comprehension).

Each task is fine-tuned for 20K iterations to capture performance across early, mid, and late training phases.
For reporting final results of our optimizer, we fix hyperparameters to $\theta = 1.4, \beta = 0.99$. For ensuring a fair comparison, we fix the learning rate to $10^{-7}$ and the smoothing parameter to $\lambda = 10^{-3}$, which is the same for MeZO.

\subsection{Additional Experimental Results}
\label{additional_results}

\begin{figure}[!ht]
  \centering
  \begin{subfigure}[t]{0.48\textwidth}
    \centering
    \includegraphics[width=\textwidth]{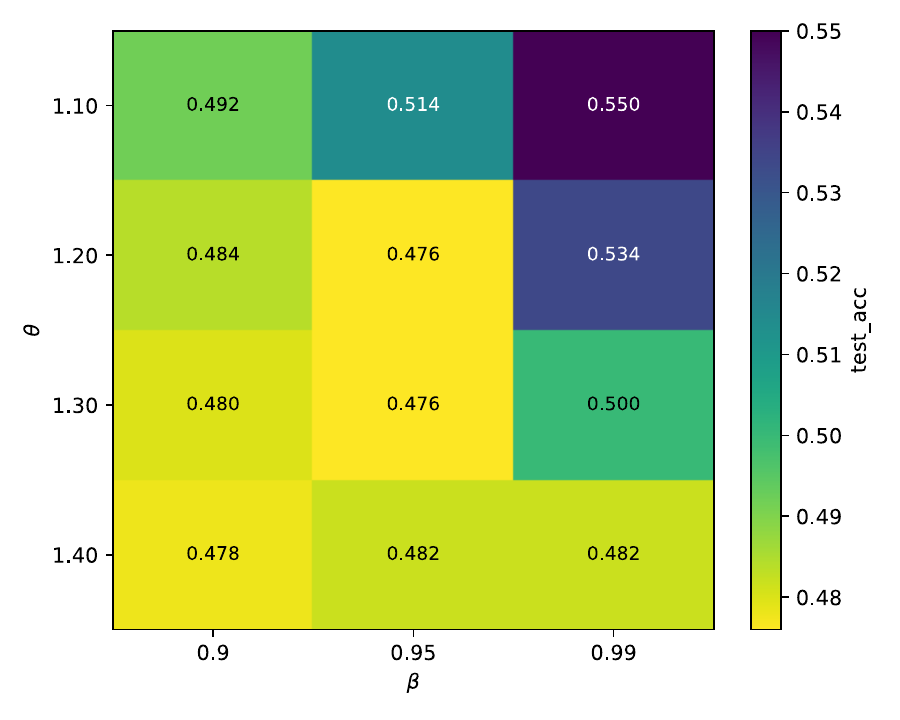}
    \caption{After 1,000 iterations. MeZO's test accuracy after 1,000 iterations is 0.474.}
    \label{fig:heatmap_trec_1k}
  \end{subfigure}\hfill
  \begin{subfigure}[t]{0.48\textwidth}
    \centering
    \includegraphics[width=\textwidth]{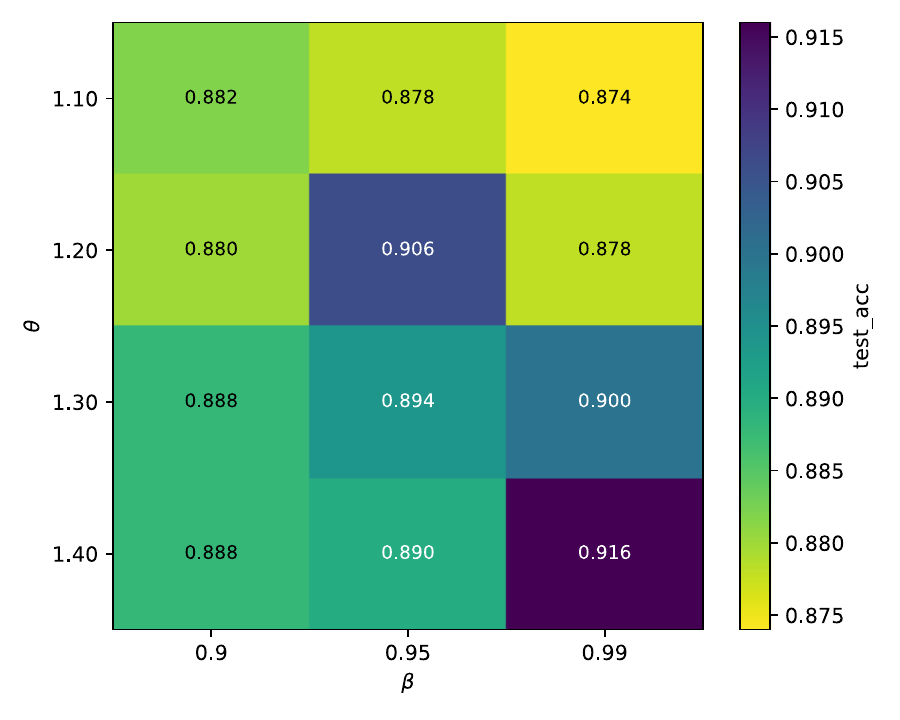}
    \caption{After 10,000 iterations. MeZO's test accuracy after 10,000 iterations is 0.89.}
    \label{fig:heatmap_trec_10k}
  \end{subfigure}
  \caption{Heatmaps of Test Accuracy of ConMeZO on TREC dataset for different $\theta$ and $\beta$ values and fixed learning rate $\eta = 10^{-6}$.}
  \label{fig:heatmaps}
\end{figure}

\begin{figure}[]
    \centering
    \begin{subfigure}[]{0.91\textwidth}
        \centering
        \includegraphics[width=\textwidth]{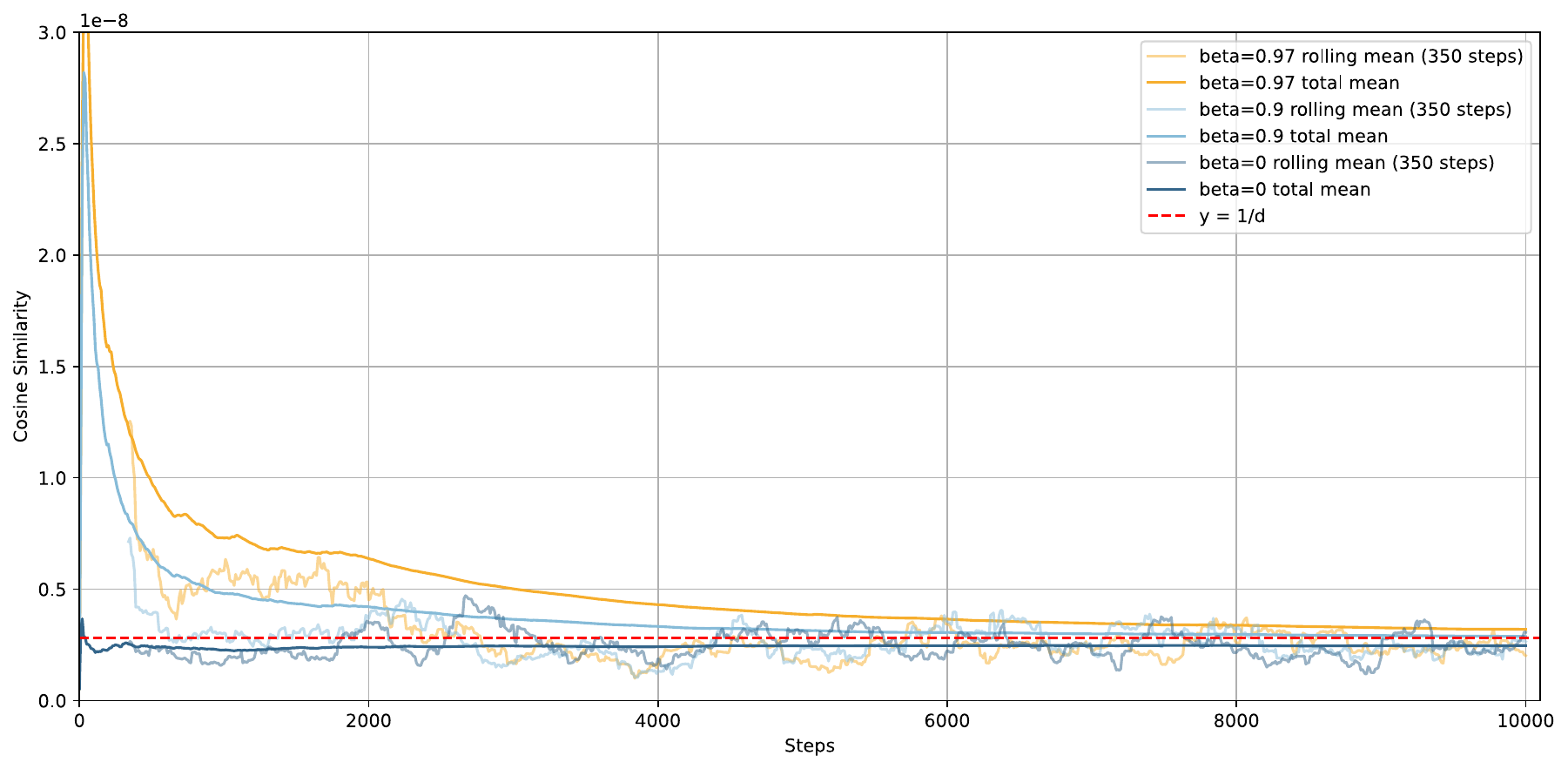}
        \caption{TREC dataset for $\theta=1.3$ and varying $\beta$.}
        \label{fig:cosine_sim}
    \end{subfigure}
    \hfill
    \begin{subfigure}[]{0.91\textwidth}
        \centering
        \includegraphics[width=\textwidth]{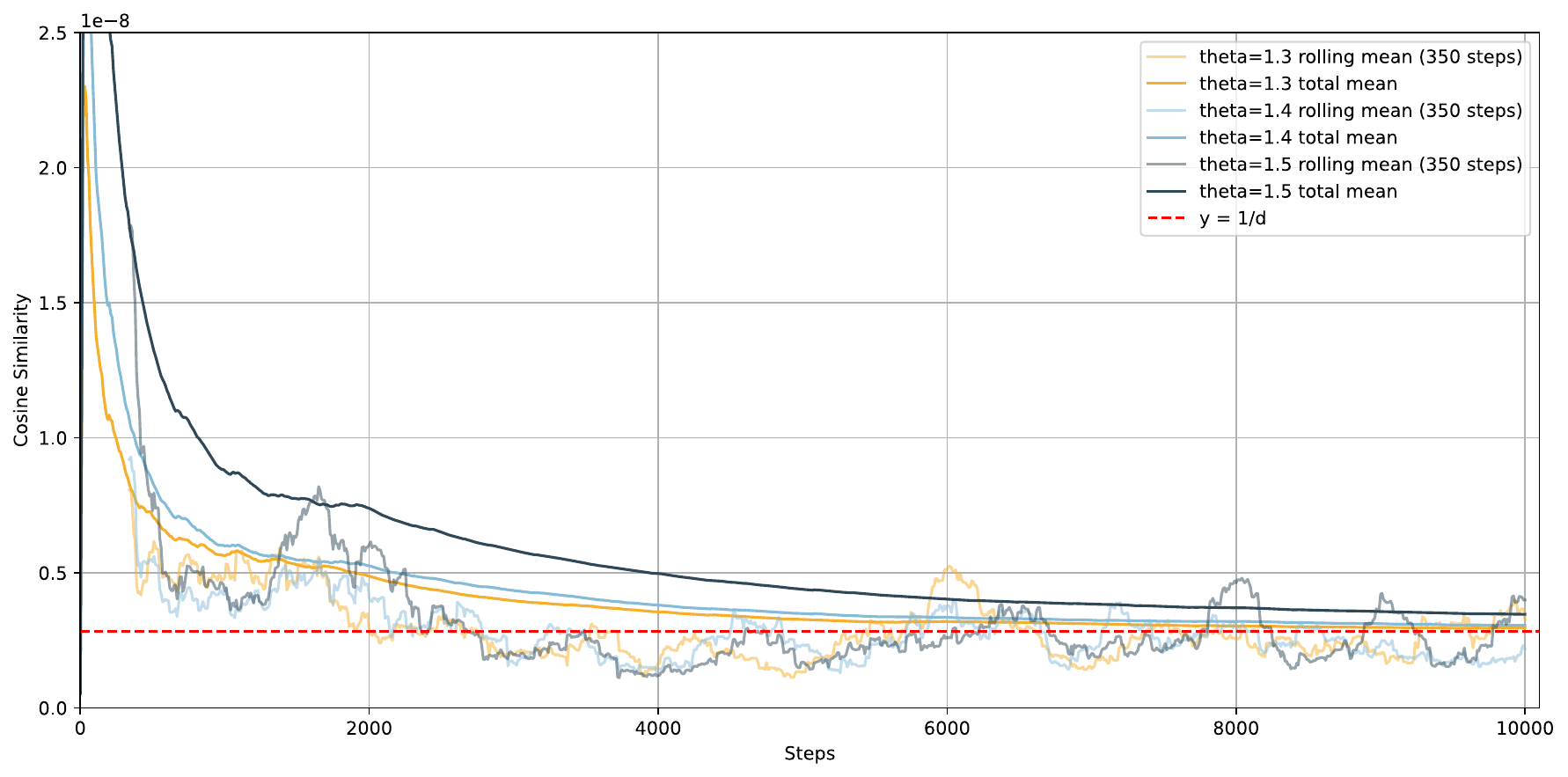}
        \caption{TREC dataset for $\beta=0.95$ and varying $\theta$.}
        \label{fig:cosine_sim2}
    \end{subfigure}
    \caption{Squared cosine similarity between the true gradient and the momentum vector during training. $\frac{1}{d}$ in red highlights the expected squared cosine similarity for a random direction.}
    \label{fig:cos}
\end{figure}

The results in Figure~\ref{fig:heatmap_trec_1k}, Figure~\ref{fig:heatmap_trec_10k}, and Figure~\ref{fig:cosine_sim2} serve as an ablation study, illustrating how different hyperparameter choices shape the optimizer's behavior and convergence dynamics.

\subparagraph{Early-Phase Convergence and Momentum Alignment.}
Momentum plays a critical role in maintaining consistent update directions during optimization. High momentum values, such as $\beta = 0.99$, help align updates with the true gradient direction, reducing variance and leading to faster and more stable convergence. As shown in the heatmaps (Figure \ref{fig:heatmap_trec_1k} and Figure \ref{fig:heatmap_trec_10k}), this effect becomes even more pronounced when combined with a small cone angle ($\theta$), while more balanced configurations yield the strongest final performance.

Our analysis of the squared cosine similarity between the momentum vector and the true gradients (Figure \ref{fig:cos}) confirms this behavior. High momentum ($\beta = 0.99$) substantially improves directional alignment, achieving up to twice the accuracy of random directions during the first 2,000 iterations. The alignment remains consistently higher throughout training, although its relative gain decreases as convergence is approached, which indicates that high momentum primarily stabilizes updates rather than dominating progress in later stages.

\subparagraph{Impact of Cone Angle: Exploration vs. Exploitation.}
The cone angle ($\theta$) balances building an accurate gradient approximation against effectively exploiting it. Larger $\theta$ values sample broader directions, increasing cosine similarity with the true gradient but reducing reliance on the estimated gradient for updates. Figure \ref{fig:cosine_sim2} shows how varying $\theta$ affects cosine similarity over iterations.

\subsection{Momentum Warm-up Details}

\label{app:warmup_details}

\textbf{Intuition.} With high momentum ($\beta = 0.99$, our default), the exponential moving average retains long memory: after 100 steps, a gradient still contributes $0.99^{100} \approx 37\%$ of the current momentum. Since early ZO estimates are noisy, assigning such large weight to initial estimates is undesirable. The warm-up starts with short memory ($\beta = 0.1$) to quickly forget early noise, then gradually increases $\beta$ so that many recent gradients contribute comparably. 

For an EMA, effective memory length scales as $1/(1-\beta)$, making the difference between $\beta=0.95$ and $\beta=0.99$ significant. Small changes near $\beta=0.99$ cause large changes in effective memory:
\begin{itemize}
    \item At $\beta=0.95$: A 100-step-old gradient contributes $0.95^{100} \approx 0.6\%$
    \item At $\beta=0.99$: A 100-step-old gradient contributes $0.99^{100} \approx 36.6\%$
\end{itemize}

\begin{figure}[h]
\centering
\includegraphics[width=0.45\linewidth]{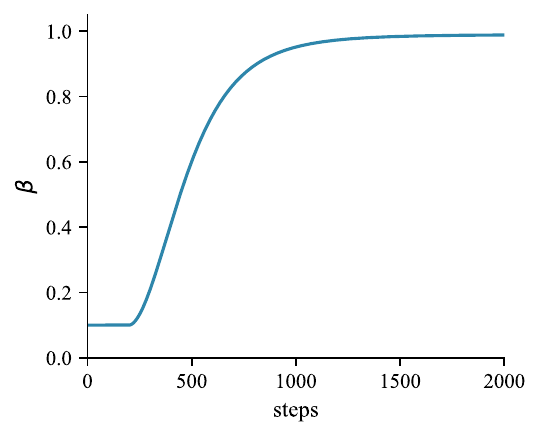}
\caption{Momentum warm-up schedule for a 20K step training run with $\beta=0.99$.}
\label{fig:warmup_schedule}
\end{figure}

Our schedule therefore increases quickly during early iterations and very slowly near the final value to match this exponential behavior. Figure~\ref{fig:warmup_schedule} visualizes the warm-up curve.

\textbf{Ablation study.} We compare ConMeZO with and without warm-up on RoBERTa-Large (5 seeds, 10K iterations).

\begin{table}[h]
\centering
\caption{Momentum warm-up Ablation study. Incorporating a warm-up phase leads to the highest average accuracy and superior performance on the majority of tasks.}
\begin{tabular}{lcccccccc}
\toprule
Method & SST-2 & SST-5 & MNLI & SNLI & RTE & TREC & Avg. \\
\midrule
MeZO & 0.928 & \textbf{0.493} & 0.697 & 0.810 & 0.739 & 0.884 & 0.759 \\
ConMeZO (no warmup) & 0.930 & 0.490 & 0.715 & 0.800 & 0.742 & \textbf{0.908} & 0.764 \\
ConMeZO (with warmup) & \textbf{0.935} & 0.489 & \textbf{0.732} & \textbf{0.819} & \textbf{0.751} & 0.900 & \textbf{0.771} \\
\bottomrule
\end{tabular}
\label{tab:warmup_ablation}
\end{table}

The warm-up yields consistent improvements across all tasks, validating its effectiveness. Note that the warm-up is a practical heuristic rather than a core algorithmic component: all theoretical results assume fixed $\beta$, and synthetic experiments (Section~\ref{sec:synthetic_experiments}) use ConMeZO without warm-up.

\subsection{Licenses}

Our evaluations are carried out on commonly-used datasets and models in the literature.

\textbf{Datasets.} GLUE \citep{wang2018glue} is designed to provide a general-purpose evaluation of language
understanding. Those adopted in our work include MNLI (inference, \cite{williams2018broad}), SST-2/5 (sentiment analysis, \cite{socher2013recursive}), SNLI (natural language inference) \citep{bowman2015large}, RTE\footnote{\url{https://paperswithcode.com/dataset/rte}} (inference), and TREC (question classification, \cite{voorhees2000building}). These datasets are released under different permissive licenses.

\textbf{RoBERTa-large.} This is a $355$M parameter model. The model checkpoint\footnote{\url{https://huggingface.co/FacebookAI/roberta-large}} is released under the MIT license.

\textbf{OPT-1.3B and OPT-13B.}
The model checkpoints\footnote{\url{https://huggingface.co/facebook/opt-1.3b}, \url{https://huggingface.co/facebook/opt-13b}} are released under a non-commercial license\footnote{\url{https://github.com/facebookresearch/metaseq/blob/main/projects/OPT/MODEL_LICENSE.md}}.
\end{document}